\documentclass[10pt,twocolumn]{article}

\usepackage[
  a4paper,
  margin=0.72in,
  top=0.78in,
  bottom=0.85in,
  columnsep=0.24in
]{geometry}

\usepackage[T1]{fontenc}
\usepackage[utf8]{inputenc}
\usepackage{lmodern}
\usepackage{microtype}

\usepackage{amsmath,amssymb,amsfonts}
\usepackage{mathtools}
\usepackage{mathrsfs}
\usepackage{bm}

\usepackage{graphicx}
\usepackage{booktabs}
\usepackage{multirow}
\usepackage{array}
\usepackage{caption}
\usepackage{subcaption}
\usepackage{float}
\usepackage{placeins}
\usepackage[rightcaption]{sidecap}

\usepackage{enumitem}
\setlist{nosep,leftmargin=*}

\usepackage{xcolor}
\definecolor{darkcyan}{rgb}{0.0, 0.55, 0.55}

\usepackage[
  colorlinks=true,
  linkcolor=darkcyan,
  citecolor=darkcyan,
  urlcolor=darkcyan
]{hyperref}

\usepackage[numbers,sort&compress]{natbib}

\usepackage{authblk}

\usepackage[compact]{titlesec}
\titlespacing*{\section}{0pt}{1.1em}{0.45em}
\titlespacing*{\subsection}{0pt}{0.9em}{0.35em}
\titlespacing*{\subsubsection}{0pt}{0.75em}{0.25em}

\titleformat{\section}
  {\large\bfseries}
  {\thesection}
  {0.6em}
  {}

\titleformat{\subsection}
  {\normalsize\bfseries}
  {\thesubsection}
  {0.6em}
  {}

\titleformat{\subsubsection}
  {\normalsize\itshape}
  {\thesubsubsection}
  {0.6em}
  {}

\usepackage{abstract}

\usepackage{xspace}
\usepackage{ifthen}
\usepackage{environ}   

\DeclareMathOperator*{\argmax}{argmax}

\makeatletter

\newcommand{\templatetype}[1]{}
\newcommand{\leadauthor}[1]{}
\newcommand{\dates}[1]{}
\newcommand{\firstpage}[1]{}
\newcommand{\dropcap}[1]{#1}
\newcommand{\acknow}[1]{}

\newcommand{\significancestatement}[1]{%
  \gdef\arxiv@significance{#1}%
}

\newcommand{\authorcontributions}[1]{%
  \gdef\arxiv@authorcontributions{#1}%
}

\newcommand{\printauthorcontributions}{%
  \arxiv@authorcontributions
}

\newcommand{\correspondingauthor}[1]{%
  \gdef\arxiv@correspondingauthor{#1}%
}

\newcommand{\keywords}[1]{%
  \gdef\arxiv@keywords{#1}%
}

\providecommand{\arxiv@significance}{}
\providecommand{\arxiv@authorcontributions}{}
\providecommand{\arxiv@correspondingauthor}{}
\providecommand{\arxiv@keywords}{}

\newboolean{shortarticle}
\newboolean{singlecolumn}
\setboolean{shortarticle}{false}
\setboolean{singlecolumn}{false}

\def\ps@firststyle{\ps@plain}

\long\gdef\arxivabstractcontent{}
\RenewEnviron{abstract}{%
  \global\let\arxivabstractcontent\BODY
}

\let\oldmaketitle\maketitle
\renewcommand{\maketitle}{%
  \twocolumn[
    \begin{@twocolumnfalse}
    \oldmaketitle
    \vspace{-1.2em}

    \begin{center}
      \begin{minipage}{0.88\textwidth}
        \small
        \noindent\textbf{Abstract.}\quad\arxivabstractcontent
      \end{minipage}

      \vspace{0.6em}
      \begin{minipage}{0.88\textwidth}
        \noindent\textbf{Keywords:}
        \arxiv@keywords

        \vspace{0.6em}
        \noindent\arxiv@correspondingauthor
      \end{minipage}
    \end{center}

    \vspace{1.2em}
    \end{@twocolumnfalse}
  ]%
}

\makeatother

\newcommand{\EagleEye}{\texttt{EagleEye}\xspace}
\newcommand{\OG}{\mathcal{Y}}

\newcommand{\IEo}{\hat{\mathcal{Y}}^+}
\newcommand{\IEu}{\hat{\mathcal{X}}^+}
\newcommand{\refe}{\mathcal{X}}
\newcommand{\test}{\mathcal{Y}}
\newcommand{\both}{\mathcal{U}}

\newcommand{\PNval}{\mathbf{\Upsilon}}
\newcommand{\phat}{\hat{p}}

\newcommand{\card}[1]{\left| #1 \right|}

\newcommand{\SIref}[1]{\hyperref[sec:#1]{\textit{#1}}}

\begin{document}

\title{Detecting Localized Density Anomalies in Multivariate Data via Coin-Flip Statistics}

\author[a]{Sebastian Springer}
\author[a]{Andre Scaffidi}
\author[b,d]{Maximilian Autenrieth}
\author[a,e]{Gabriella Contardo}
\author[a,1]{Alessandro Laio}
\author[a,d]{Roberto Trotta}
\author[c]{Heikki Haario}

\affil[a]{Scuola Internazionale Superiore di Studi Avanzati (SISSA), Trieste, Italy}
\affil[b]{University of Cambridge, Cambridge, United Kingdom}
\affil[c]{LUT University, Lappeenranta, Finland}
\affil[d]{Imperial College London, London, United Kingdom}
\affil[e]{University of Nova Gorica, Nova Gorica, Slovenia}

\leadauthor{Springer}

\significancestatement{
Scientific discoveries often hinge on subtle, localized differences in high-dimensional data, such as rare-event signatures, regime shifts, or mismatches between observations and simulations.   Many standard tools are limited to detecting either global differences between datasets or isolated outliers, offering little insight into where discrepancies arise. We present \EagleEye, a deterministic, training-free two-sample method that performs local comparisons and identifies statistically principled, interpretable regions where one is systematically over- or under-represented relative to the other. \EagleEye returns clusters of anomalous points and neighbourhoods, enabling targeted, in-depth inspection. We demonstrate its performance and versatility on synthetic benchmarks, new-physics searches with multiple background models, and climate reanalysis fields.
}

\authorcontributions{
\textbf{Conceptualization:} S.S., A.L., H.H. conceived the study and defined the scope of \EagleEye.
\textbf{Methodology:} All authors; anomaly-score framework: S.S., H.H., A.L.; over/under-density formalization: M.A., G.C.; IDE and rep\^echage: S.S., A.L.; background injection: M.A., R.T.; purity estimation: M.A., A.S., G.C.
\textbf{Software:} S.S., A.S.
\textbf{Investigation:} Collider application: A.S. (with G.C., R.T., M.A., SS, AL); climate application: S.S. (with A.L., H.H.); numerical experiments/refinements: S.S., A.S., M.A., G.C.
\textbf{Visualization:} S.S., A.S., G.C. (with input from all authors).
\textbf{Writing:} All authors.
}

\correspondingauthor{\textsuperscript{1} E-mail: laio@sissa.it}


\begin{abstract}
Detecting localized differences between two samples is a central task in scientific data analysis, required for the identification of signal events, regime changes, or model mismatch. 
We introduce \EagleEye, a  method that pinpoints \emph{local} over- and under-densities in multivariate feature spaces.
\EagleEye assigns each point an anomaly score by encoding its ordered $k$-nearest-neighbour list as a binary membership sequence and testing whether the cumulative number of successes in this sequence is consistent with a binomial (coin-flipping) null model.
In the presence of a genuine local anomaly, neighbours will preferentially belong to one of the two datasts, yielding an excess of ``successes'' relative to the binomial null model.
These local, pointwise detections are consolidated into interpretable anomaly sets through a deterministic refinement procedure that can also estimate the irreducible background and local density anomaly purity. 
We demonstrate \EagleEye's efficacy  in three scenarios. We first consider an artificial data example with known localized over- and under-densities. Second, we demonstrate how \EagleEye may be used for new physics searches at particle collider experiments in the presence of systematic background modelling differences. Finally, we conduct a climate analysis study that reveals localized changes in spatiotemporal temperature-pattern recurrence. 
\end{abstract}

\dates{This manuscript was compiled on \today}

\maketitle
\thispagestyle{plain}

\dropcap{M}any scientific pipelines involve comparing two collections of high-dimensional observations: a \emph{reference} sample representing expected behaviour (e.g., a baseline simulation, data from a trusted historical period, or other controlled conditions) and a \emph{test} sample that may or may not differ (e.g., an alternative simulation, a different temporal regime, or an unverified process) \cite{chandola2009anomaly,Ruff2021}. In many settings, the two samples do not differ globally, but only \emph{locally}. That is, in relatively small regions of feature space where the test distribution exhibits an excess or deficit relative to the reference. Detecting such localized density anomalies  is central to a wide range of scientific tasks, from rare-event discovery to change detection under evolving experimental or environmental conditions. Many two-sample anomaly detection approaches are powerful for rejecting distributional equality, but provide limited guidance about \emph{where} and to \textit{what extent} the discrepancies occur \cite{Gretton2012,Szekely2013,Friedman1979,Schilling1986}. Others \cite{lemos2024pqmass} rely on first partitioning the space before aggregating into a final score or p-value. More generally, detecting density anomalies in high-dimensional data without relying on complex models or intensive computation remains challenging. This difficulty is further amplified when the reference distribution is not uniquely defined: in many scientific applications, the reference distribution may be generated under multiple plausible models or simulation assumptions, requiring assessment of whether observed discrepancies are robust to the choice of reference \cite{TheNCEPNCAR40YearReanalysisProject,Berge_2007}. Local comparisons offer a natural way to overcome the limitations of global two-sample testing. Existing approaches include kernel-based local tests and density-difference estimators \cite{rasmussen2006gaussian,duvenaud2014cookbook,duong2013local}, and neural methods that learn density ratios or classifiers for localized discrepancy detection \cite{DAgnolo2021LearningMultivariateNewPhysics,logistic_boosting,Sugiyama_Suzuki_Kanamori_2012,CWoLa}. In high dimensions these approaches face practical obstacles: kernel methods demand careful bandwidth selection because data typically lie on non-linear, low-dimensional manifolds, while neural methods can suffer from suppressed gradient updates when the anomaly occupies an exceedingly small region, causing classifiers to collapse to trivial solutions \cite{vanishing_grads,279181}. 

To address these challenges, we introduce \EagleEye, which avoids explicit density estimation and bandwidth tuning by assigning each data point an anomaly score computed from the ordered ranks of its nearest-neighbours. Computationally, the procedure is dominated by the $k$NN search and therefore inherits the scaling of the chosen neighbour-search algorithm, while remaining readily parallelisable.   

The \EagleEye method links density anomaly detection to a simple and intuitive process: flipping a  coin. Consider two datasets, one representing the \textit{reference} data, and another that may contain localized over or under-densities of points i.e the \textit{test} data. For each point in the \textit{test} data, we examine its nearest neighbours in both datasets. If the two datasets are generated from the same underlying probability density distribution, these neighbours will constitute a locally exchangeable mix of both datasets, much like expecting an equal number of heads and tails when flipping a fair coin many times. 
On the other hand, if a local ensemble of points from the \textit{test} significantly deviates in concentration from the reference data, points belonging to the \textit{test} set will be  over- or under-represented in its neighbourhood. This constitutes what is called a \textit{density anomaly}. The analogy here is similar to flipping a coin multiple times and obtaining an unexpectedly large number of heads or tails, which might suggest that the coin is biased. \EagleEye quantifies this imbalance by ``counting" the heads and tails in the local neighbourhood of each test point. By carrying out a statistical hypothesis test on whether the observed mix of neighbours significantly deviates from what is expected under the hypothesis of random chance (akin to the binomial distribution of coin flips), \EagleEye  can therefore flag potentially  anomalous points. Importantly, this test can be performed \textit{individually} for \textit{each} point. This allows the identification of \emph{local density anomalies} in a statistically principled manner.


Whilst based on a fundamentally simple premise, the \EagleEye method consists of several novel steps which allow for a complete study of local distributional discrepancies. \EagleEye natively handles data imbalance (differing cardinality) between the \textit{reference} and \textit{test} sets, since the statistical measure used to construct scores is parametrically well defined. The method is also deterministic, not relying on any stochastic optimisation or training procedure, facilitating interpretability.    
The procedure is symmetric: although we  label one sample as \textit{test} and the other as \textit{reference} for convenience, we run the analysis in both directions, yielding a complete picture of localized excesses and deficits in each sample relative to the other. Note that \EagleEye is not an outlier detection method in that it will not detect a single outlier point as anomalous. Unlike classifier-based tests, \EagleEye\ requires no training phase or sophisticated hyperparameter tuning.  Crucially, its statistic is built from neighbour ranks,  which are largely invariant when the data lie on a low-dimensional manifold embedded in a higher-dimensional space; fixed-bandwidth kernel estimators, by contrast, smear probability mass into empty directions and can therefore bias density estimates in such settings \cite{kde_curved_domains}.

We next summarize the \EagleEye\ workflow and its outputs in \emph{The EagleEye framework}, followed by three distinct scientific applications in \emph{Results} that demonstrate how \EagleEye\ operates across varying data contexts. Specifically, we present a controlled synthetic walkthrough, a collider physics-inspired study highlighting robustness to reference-model misspecification, and a climate analysis revealing localized changes in spatiotemporal temperature-pattern recurrence.

\section*{The EagleEye framework}
\label{sec:EE_framework}
\EagleEye addresses a common two-sample problem: given a \emph{reference} sample \(\refe = \{X_1, X_2, \dots, X_{n_\mathcal{X}}\}\) and a \emph{test} sample \(\test = \{Y_1, Y_2, \dots, Y_{n_\mathcal{Y}}\}\), each drawn from $d$-variate distributions, it aims to identify \emph{localized density anomalies}, that is, regions where $\mathcal{Y}$ is significantly over- or under-represented relative to $\mathcal{X}$. The procedure is symmetric with respect to the two samples and is run in both directions to capture localized excesses and deficits. 

At its core, \EagleEye turns local neighbourhood composition into a calibrated pointwise score. For each point, we compute its nearest neighbours in  $\refe\cup\test$, up to a user defined maximal neighbourhood rank $K_M$, and examine the ordered sequence of its neighbors' memberships (as belonging to $\mathcal{X}$ or $\mathcal{Y}$). The membership of the $l$-th neighbor of data point $i$, which we denote by  $b^l_i$, takes the value 1 if the $l$-th neighbor belongs to the same sample, and 0 otherwise.  Under the null hypothesis that the two samples are locally exchangeable, the membership sequence behaves like independent coin tosses with a probability determined by the global mixture proportion. For each data point this allows for the calculation of an anomaly score based on the most significant deviation from the expected ``coin-flip" behavior. This anomaly score, which we denote $\PNval_i$, is compared against a threshold  $\PNval^*$, calibrated to achieve an exceedance probability of $p_{\mathrm{ext}}$ as described in \textit{Materials and Methods}.  Observations with $\PNval_i > \PNval^*$ are flagged as putatively anomalous. This \emph{flagged} set can be interpreted as an anomaly map comprised of an anomalous core with a surrounding ``halo'' of points that are adjacent to the density anomaly, but have also exceed the threshold when sufficiently large neighbourhoods were considered. We then refine and catalogue these putative detections into localized, interpretable anomaly sets through two algorithmic steps. \emph{Iterative Density Equalization (IDE)} iteratively selects the highest-scoring candidate in the flagged subset, removes it (a process we call \emph{pruning}), and recalculates the anomaly score for the remaining points. This process continues until no points remain flagged as anomalous. The resulting set of "pruned" points are considered representative of the local density excesses. The final step of the algorithm, which we call \emph{repêchage}, sharpens the flagged set by recalibrating a cluster-level threshold using the pruned points from the IDE stage. We first cluster the flagged points and compute the empirical distribution of anomaly scores $\PNval_i$ for those that were pruned by IDE. We then set the local threshold to a small quantile of this distribution (by default $q=0.01$). Points are declared anomalous if their $\PNval_i$ exceeds the corresponding cluster-specific threshold. Finally, within each mode, \EagleEye estimates how many of the flagged points are expected to contribute to an 'irreducible background',  providing a per-anomaly estimate of signal-to-noise and signal purity.

\begin{figure}[h!]
    \centering
    \includegraphics[width=\linewidth]{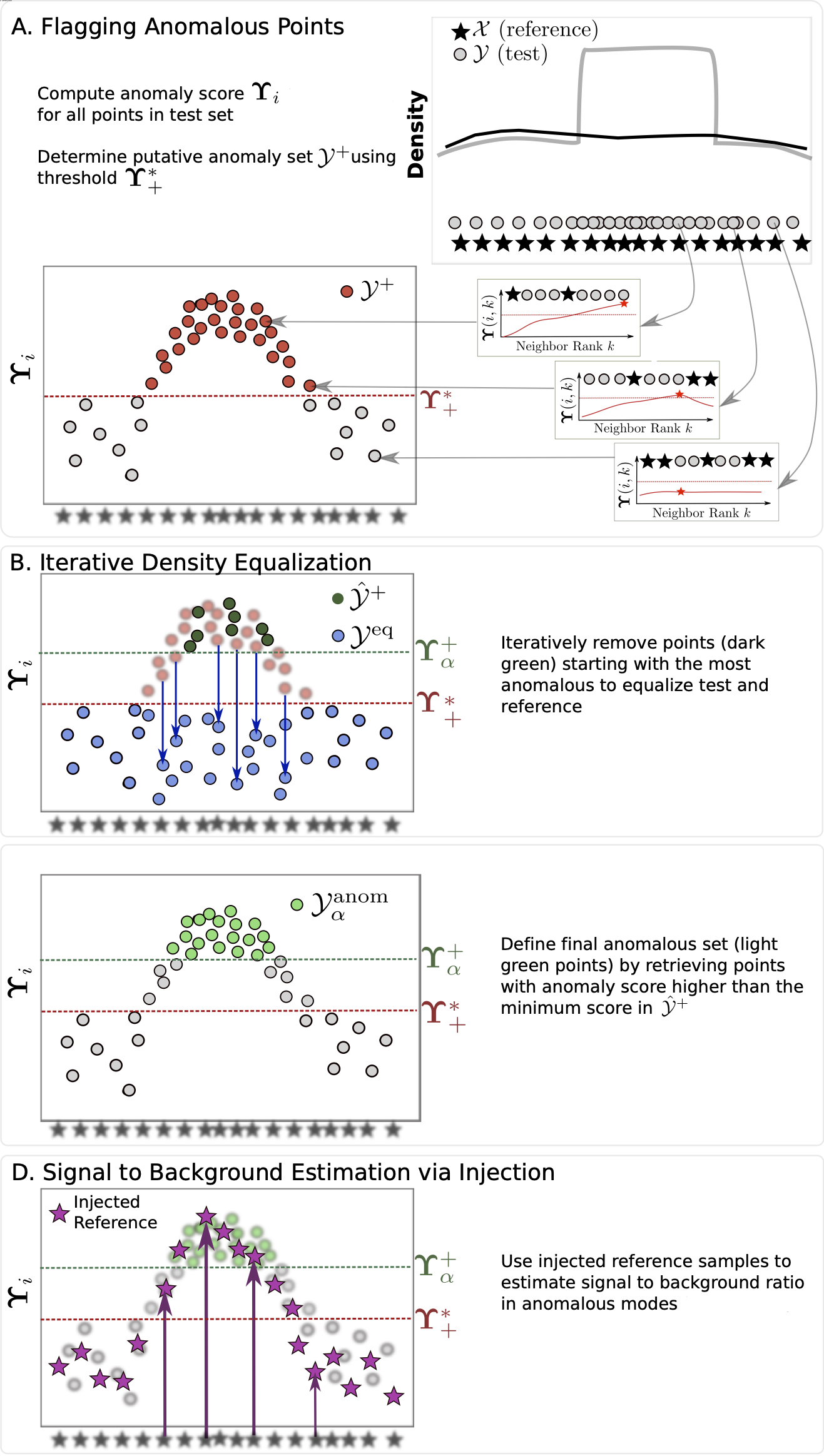}
    \caption{Illustrations demonstrating how localized density differences between two datasets are detected with \EagleEye. The method scores each point by how its nearby neighbours are distributed across the two samples, flags candidate anomalies, and refines them into clustered anomaly sets. See \textit{Materials and Methods} for details.}
    \label{fig:fig01}
\end{figure}

The full technical details of EagleEye, including the anomaly score definition and algorithmic stages, are provided in  \textit{Materials and Methods}; Fig.~\ref{fig:fig01} provides a schematic overview of the method’s main stages.

\section{Results}
\label{sec:Results}
We demonstrate the applicability of \EagleEye across three scenarios, showcasing its versatility, adaptability, and interpretability in detecting localized density anomalies.  First, we provide a synthetic case study to illustrate how the method performs on simulated data to reveal localized anomalies, and offer a step-by-step walkthrough of the workflow. Next, we apply \EagleEye on simulated high-energy physics data to perform anomaly detection in the presence of two different reference samples. Finally, we explore its use on climate  data, detecting emergent shifts in spatial and temporal temperature patterns that are not captured by traditional analyses. 

\subsection{Gaussian anomalies in a uniform background}
\label{sec:synthetic_uniform_gaussians}

\begin{figure*}[t!]
\centering
\includegraphics[width=\textwidth]{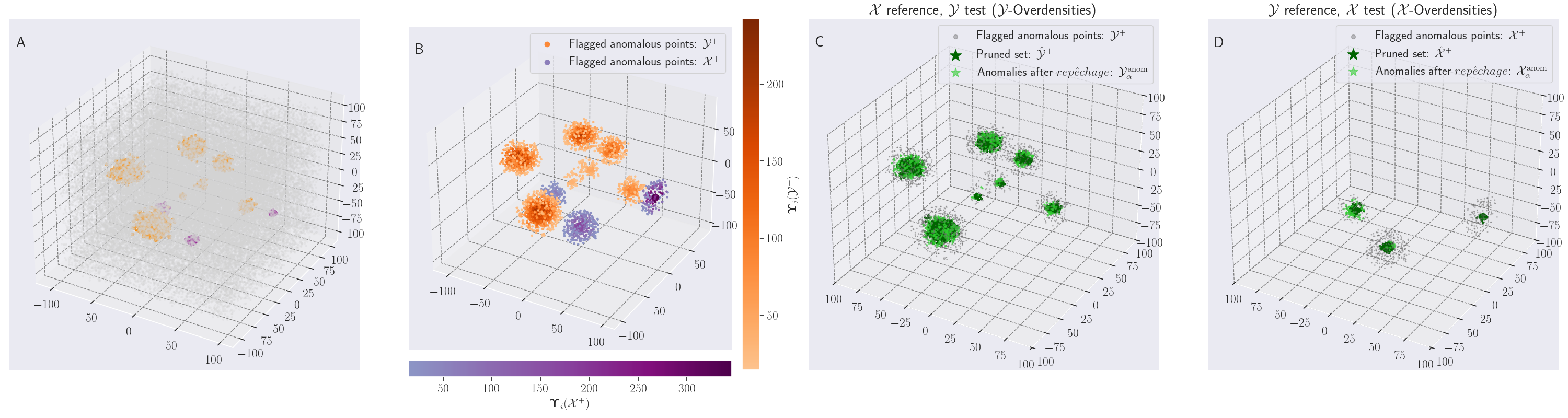}
\caption{ Detection of localized density anomalies within a uniform background with \EagleEye. \textbf{(A)} Ground-truth in feature space (not shown to \EagleEye): injected overdensities (orange) and corresponding underdensities (violet). \textbf{(B)} Points initially flagged as anomalous in the test set ($\mathcal{Y}$; warm colors) and in the reference set ($\mathcal{X}$; cool colors). \textbf{(C,D)} Refinement of detections after iterative density equalization (IDE; dark green) and multimodal rep\^echage (light green), shown separately for overdensities (left) and underdensities (right).}
\label{fig:synthetic_main}
\end{figure*}

\paragraph{Setup.}
As a first illustrative example we construct two synthetic three-dimensional datasets, each with $5\times10^{4}$ points. The test sample \(\mathcal{Y}\) comprises a uniform background plus seven isotropic Gaussian components (``\(\mathcal{Y}\)-overdensities''), each with distinct means and variances chosen so that the probability of any  components producing samples occupying the same region is negligible. The reference sample $\mathcal{X}$ contains a uniform background plus three isotropic Gaussian components (``$\mathcal{X}$-overdensities''). Because $\mathcal{X}$ has higher local density in these regions, $\test$ exhibits a corresponding local \emph{underdensity} around each $\mathcal{X}$-overdensity.
 This controlled construction provides ground truth for both over- and under-represented regions, enabling an illustration of the \EagleEye\ workflow. We provide details about the data generation in \ref{sec:synth_data_details}.

\paragraph{Workflow.}
We run \EagleEye\ symmetrically in both directions ($\mathcal{X}$ as reference versus $\mathcal{Y}$ as test, and vice versa) to capture overdensities and underdensities as separate passes. 

The analysis proceeds in three stages: (i) \emph{flagging} candidate anomalous points; (ii) \emph{iterative density equalization} (IDE) to identify representatives of local density excesses; and (iii) \emph{repêchage}, which clusters flagged candidates into localized anomaly sets and sharpens their boundaries using an IDE-driven, cluster-specific threshold.

\paragraph{Outputs.}
Fig.~\ref{fig:synthetic_main} depicts the sets obtained at the end of the first three stages of the pipeline outlined in Materials and Methods.
Panel~A shows the ground truth. Panel~B highlights that the flagging step recovers points sampled from the Gaussian components, while also marking some nearby background points whose neighbourhoods are influenced by the same Gaussians. IDE (Panels~C,D; dark green) constructs a pruned set that is representative of the local density excess in each region, thereby excluding halo points whose anomalous scores arise simply from proximity to a true component. The final rep\^echage sets (light green) capture the bulk of points sampled from the anomalous Gaussian components while minimizing inclusion of nearby, non-anomolous background points, yielding interpretable anomaly clusters and the final objective of the \EagleEye method. We provide further details regarding this demonstrative example in \ref{sec:synth_data_details}. Specifically, we display component-wise recovery in Table~\ref{tab:S1} and an extended figure, including score distributions, in Fig.~\ref{fig:supp_SI1}.

\paragraph{Additional synthetic tests.}
Beyond this illustrative walkthrough, \ref{sec:SI1g_manifold} considers more challenging synthetic regimes. We study the performance on data lying on a nonlinear 10-dimensional manifold embedded in 100 dimensions, including background-only comparisons to assess false-positive control, as well as signal-injection experiments to quantify the number of correctly recovered samples in the detected anomalies. We also provide quantitative comparisons with commonly used two-sample classifier-based tests, including a multilayer perceptron (MLP) and an RBF-kernel classifier (\ref{sec:two_sample_compt_intro}). On the 10D-in-100D benchmark, \EagleEye’s pointwise anomaly scores achieve stronger ROC performance than the MLP. In an extreme “needle-in-a-haystack” regime (a highly concentrated anomaly comprising very few points), the MLP fails to recover the anomaly, whereas \EagleEye detects all points constituting the over-density. For a structured spiral anomaly, the RBF-kernel baseline yields diffuse detections, while \EagleEye localizes the full structure and recovers all anomalous points. Notably, both classifier baselines require heuristic threshold selection, whereas \EagleEye relies on statistically principled and calibrated anomaly scores.
We report sensitivity analyses with respect to hyperparameters for \EagleEye in \ref{sec:SI1} and \ref{sec:SI1g_manifold}.


\begin{figure*}[t]
    \centering
    \includegraphics[width=.7\textwidth]{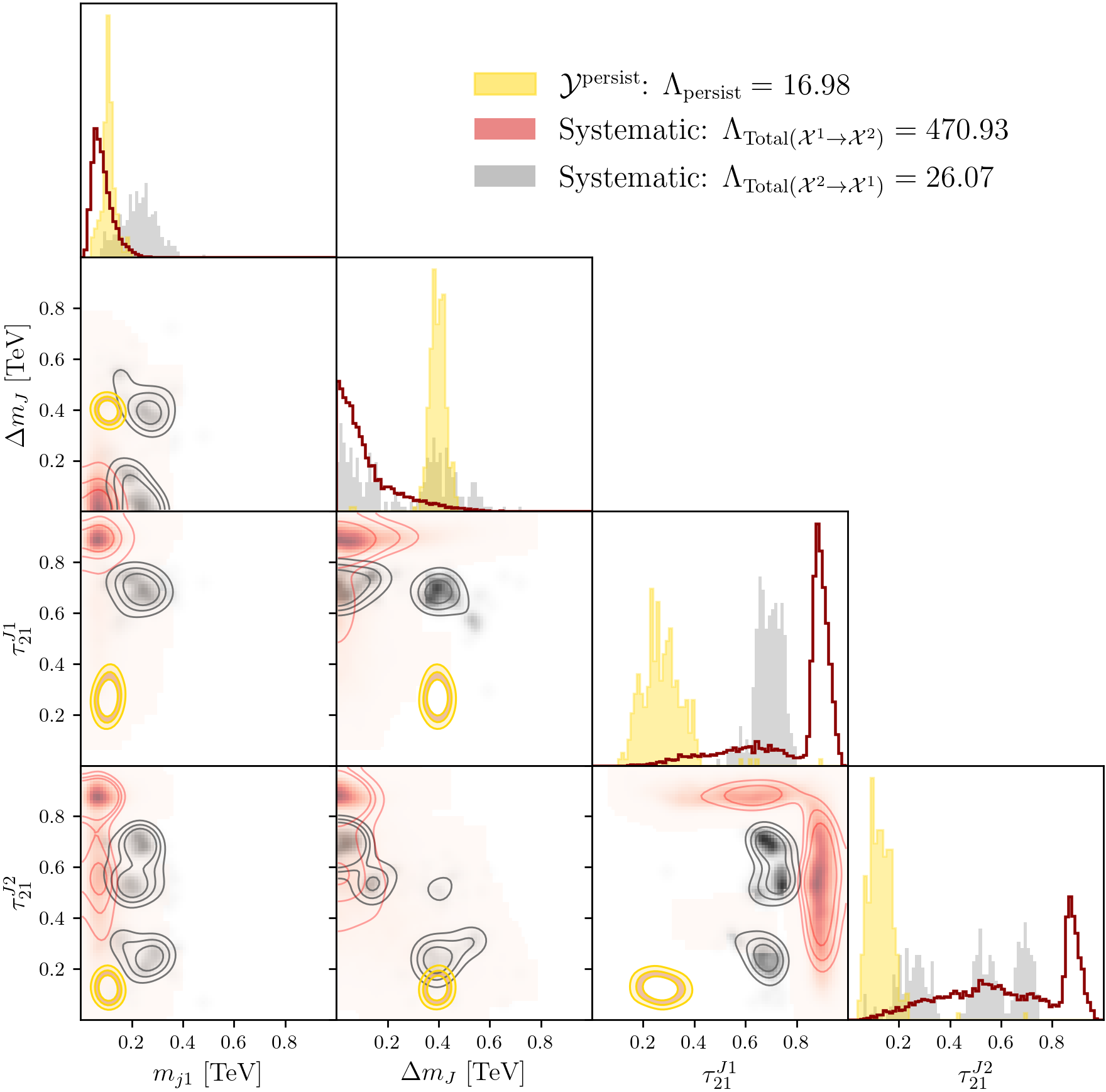}
\caption{
\textbf{Systematics mapping and identification of persistent anomalies in the presence of multiple background models in  simulated data at the Large Hadron Collider.}\newline \textbf{Red and grey contours: } 
Multivariate systematics map showing  discrepancies between two background models 
$\refe_1$ (\textsc{Pythia}) and $\refe_2$ (\textsc{data driven}) using \EagleEye, in the 4-dimensional feature space $(m_{j1},\,\Delta m_J,\,\tau_{21}^{J_1},\,\tau_{21}^{J_2})$ (see SI2 for feature details).
Red regions/histograms denote overdensities in $\refe^1$ compared to $\refe^2$, while grey regions are overdensities in $\refe^2$ compared to $\refe^1$, with the shading and outermost contour representinng the 68\% highest density regions of the points, with full marginal density displayed in the diagonals. The two background models are free from discrepancies in the white regions, where an anomaly search is unimpeded by background modelling systematics. 
\textbf{Yellow contours: }
The persistent anomaly search described in the text yields the yellow regions/histogram, which lies in a systematics-free part of feature space. 
The estimated $z$-score for the persistent anomaly (calculated with the values in Tab.~\ref{tab:LHC_cluster_summary} using Eq.\ref{eq:lambda_persist_pointwise}) is
${\Lambda}^\text{persist}=16.98$, compared with true value (after unblinding)
$\Lambda^\text{persist}_{\text{True}}=16.13$, demonstrating good agreement
the estimate. This example shows how \EagleEye can be used to identify density anomalies when different background models are adopted as multiple reference sets, all whilst mapping any discrepancies between the background models. 
}
\label{fig:LHC_locations}
\end{figure*}
\subsection{Persistent anomalies across multiple background models:  
New physics searches and systematics mapping at the Large Hadron Collider (LHC)}
\label{sec:LHC_results}

In this section, we demonstrate  how \EagleEye addresses two distinct but linked scientific problems: (i) \textit{Quantification of discrepancies among different background models}, and (ii) \textit{robust anomaly detection in the presence of multiple discrepant background models}.


\paragraph{Setup.}
In collider experiments, searches for new physics beyond the Standard Model require identifying 
overdensities with respect to a background distribution.  In practice, the background is not perfectly known: different Monte Carlo generators, detector simulations, and theory predictions produce multiple plausible background models whose discrepancies constitute dominant systematic uncertainties, referred to as \textit{systematics}. Current approaches for quantifying such systematics  include shape, or normalisation-based nuisance parameter treatments \cite{CRANMER_2006},   low-dimensional ``MC-closure’’ residual corrections \cite{CMScollaboration_2011,Fleck_2013} and sideband-based interpolation methods \cite{Cowan:2010js,CWoLa}.

\EagleEye offers a complementary approach. We consider a resonant anomaly search based on the LHC Olympics R\&D dataset \cite{zenodo_rnd}.  The dataset is built from realistic simulations of standard-model dijet background events, generated with \textsc{Pythia}~8 \cite{pythia} and passed through a detector simulation using \textsc{Delphes}~3.4.1 \cite{delphes}.  To mimic the presence of new physics, a small number of signal events are injected on top of this background.  These signal events originate from a heavy resonance, corresponding to a hypothetical new particle that decays into detectable final-state particles, and therefore manifests as an overdensity anomaly.  The resulting mixture of background and injected signal is the test dataset, $\test$.

In order to mimic a real analysis with multiple background models, we adopt two different reference sets, $R=[\refe^1,\refe^2]$. The first consists of simulated QCD dijet events produced with the same \textsc{Pythia}~8 hadronic simulation, but reconstructed using different detector simulation (\textsc{Delphes}) hyperparameters, yielding a background that slightly deviate from that in $\test$. The second is obtained using an interpolation approach following the framework of \cite{cathode2022}, in which the background is inferred under weakly supervised assumptions directly from data (in this case $\test$) rather than generated from simulation. All events are reduced to the four-dimensional feature vector defined in Eq.~(\ref{eqn:LHC_features}), with additional feature engineering following standard LHC practice\footnote{We note that work has also been done investigating optimal representations for anomaly detection tasks \cite{Dillon_2024}. Using \EagleEye with optimal representations is an interesting prospect which we leave for future work}.  The resulting test dataset contains approximately $0.6\%$ injected signal, comparable to the signal-to-background composition of Black Box~1 in the LHC Olympics challenge \cite{Kasieczka_2021}. Full details of the dataset construction, background definitions, feature engineering and sample sizes are provided in \ref{sec:lhc_features}.

\paragraph{Background models comparison: construction of the systematics map.}
To characterise density differences between background models, and thus solve problem (i) above, we construct a \emph{systematics map} by comparing the two reference (i.e., background models) samples. This aims to identify regions of feature space where the background models differ, and thus should be considered unreliable domains for anomaly searches. The systematics maps are obtained with $p_\text{ext}=10^{-5}$, $K_M=500$.  We first run \EagleEye using $\refe^1$ as reference and $\refe^2$ as test set, in order to identify overdensities in $\refe^2$ with respect to $\refe^1$, and we denote the ensuing clusters  $\alpha(\refe^1 \rightarrow \refe^2)$. We identify two clusters of anomalous points, the union of which is shown as red contours in Fig.~\ref{fig:LHC_locations}.  We then reverse the role of reference and test, obtaining overdensities in $\refe^1$ with respect to $\refe^2$, labelled $\alpha(\refe^2 \rightarrow \refe^1)$. Here too we observe two clusters that are displayed as grey contours in Fig.~\ref{fig:LHC_locations}.  Together, the red and grey contours constitute regions of feature space where the background models differ. We introduce the estimated statistic $\Lambda_{\mathrm{Total}(\refe^i \rightarrow \refe^j)}$,  as an approximate $z$-score in the large-background regime for the union of both clusters. It quantifies the strength of the total discrepancy between the two background models (see ~\eqref{eqn:s/b} in Materials and Methods for a precise definition per cluster $\alpha$). We find $\Lambda_{\textrm{Total}(\refe^1 \rightarrow \refe^2)} = 470.93$ and $\Lambda_{\textrm{Total}(\refe^2 \rightarrow \refe^1)} = 26.07$, indicating highly significant discrepancies.  Regions of feature space not belonging to the clusters are domains in which both background models are in agreement (and therefore, assumed reliable). Later, they can be used to cross-check that any identified anomaly occurs in a region where background mismodelling is unlikely. 

\begin{table*}[t]
\centering
\caption{%
Summary of overdensity clusters identified by \EagleEye across two reference sets $\refe^1$ and $\refe^2$, ordered by their estimated $z$-score, $\Lambda_{\alpha(\refe^j\rightarrow\test)}$. The numerosity of each cluster is given by $n_\alpha$. The last column quantifies the overlap between each significant anomaly cluster that is detected in the test set relative to the reference set $\refe^1$ and the only significant anomaly cluster detected in the test set relative to $\refe^2$. Bold lines indicate the persistent anomaly identified.
}
\label{tab:LHC_cluster_summary}
\begin{tabular}{lrrr}
\toprule
Anomaly cluster $\mathcal{Y}^{\text{anom}}_{\alpha}$  & $n_\alpha$ & $\Lambda_{\alpha(\refe^j\rightarrow\test)}$ &  Overlap with ${\mathcal{Y}^{\text{anom}}_{0}(\refe^2)}$    \\
\midrule
$\mathcal{Y}^{\text{anom}}_{0}(\refe^1)$  & 4799 & 203.51   & 0 \%  \\
{$\boldsymbol{\mathcal{Y}^{\text{anom}}_{1}(\refe^1)}$}  
& \textbf{164}  
& \textbf{25.98}  
& \textbf{55.90\%}  
\\
{$\boldsymbol{\mathcal{Y}^{\text{anom}}_{0}(\refe^2)}$}  
& \textbf{285}  
& \textbf{13.31}  
&  \textbf{100\%} 
\\
\bottomrule
\end{tabular}
\end{table*}

\paragraph{Persistent anomaly in the presence of multiple backgrounds.}

We now illustrate how \EagleEye can be used to identify density anomalies in the presence of multiple background models, thus addressing problem (ii) above. \EagleEye is run twice, once with $(\refe^1\rightarrow\test)$ and once with $(\refe^2\rightarrow\test)$, in order to identify potential anomaly clusters in the test set with respect to either background model\footnote{We do not run the reverse comparison, as in this specific physics setting any genuine anomaly will only appear as an overdensity \cite{Choudalakis:2011qn}.}. 
A candidate anomaly is interpreted as signal only if it \emph{persists} with respect to both background models. Persistence is determined as follows: each anomolous cluster $\alpha$ must exhibit a sufficiently large standardized discrepancy $\Lambda_{\alpha(\refe^j\rightarrow\test)}$; and (2) for every cluster $\alpha$ passing criterion (1), $\exists $ at least one other cluster from each comparison $\refe^j\rightarrow\test$ exhibiting non-zero overlap. Criteria (1) and (2) also extend to the case of more than two references. The persistent anomaly is then the union of the anomalies that fulfil the above criteria\footnote{Anomalies in regions where the alternative background estimates disagree require further study, and may be missed by our criteria for persistence. In practice, LHC analyses typically avoid or tightly control such systematic-dominated regions. We leave detailed treatment of these cases for future work. Secondly, it is possible that two clusters may share one or few points by chance. If this is observed, we recommend that association be determined by the clusters that exibit maximal overlap.   }.

For each analysis $\refe^j\rightarrow\test$, \EagleEye was run with the same hyperparameters used to construct the systematics map. 
All identified repêchage clusters were observed to be stable under variations of the hyperparameter used in the clustering algorithm detailed in \textit{Materials and Methods}. The resulting anomalous overdensity clusters are summarised in Table~\ref{tab:LHC_cluster_summary}, using the notation $\mathcal{Y}^{\text{anom}}_{\alpha}(\refe^j)$ to denote the $\alpha$-th overdensity cluster
in the test sample $\test$  relative to background reference $\refe^j$ after repêchage, ranked according to their estimated $z$-score, $\Lambda_{\alpha\left(\mathcal{X}^j \rightarrow \mathcal{Y}\right)}$.

Among the 3 overdensity clusters identified in $\test$ (with respect to either background), all satisfy the persistent anomaly criterion (1). Since $(\refe^2\rightarrow\test)$ has only one cluster, criterion (2) is verified by evaluating the overlap of each of the two clusters for $(\refe^1\rightarrow\test)$ with  ${\mathcal{Y}^{\text{anom}}_{0}(\refe^2)}$. The only non-zero overlap is with 
$\mathcal{Y}^{\text{anom}}_{1}(\refe^1)$, thus indicating the existence of a persistent anomaly, for which we define as the union
\[
\mathcal{Y}^{\text{persist}}
\;:=\;
\mathcal{Y}^{\text{anom}}_{1}(\refe^1)\,\cup\,\mathcal{Y}^{\text{anom}}_{0}(\refe^2).
\]
We quantify the $z$-score of $\mathcal{Y}^{\text{persist}}$, $\Lambda^\text{persist}$,  using a weighted combination of the estimates $\Lambda_{\alpha\left(\mathcal{X}^j \rightarrow \mathcal{Y}\right)}$ (see \ref{sec:multi_ref_significance}, Eq.~\ref{eqn:lambda_peersist}), obtaining ${\Lambda}^\text{persist}=16.98$.
After unblinding the data labels, the true value of the $z$-score within $\mathcal{Y}^{\text{persist}}$ is found to be $\Lambda^\text{persist}_\text{True}=16.13$, demonstrating good agreement with our estimate. In this context, the persistent anomaly is interpreted as evidence for a new physical process,  or 'bump' as colloquially known in the particle physics community, whose density field is represented with the yellow contours in Fig.~\ref{fig:LHC_locations}.  Applying criterion (2) identifies the cluster $\mathcal{Y}^{\text{anom}}_{0}(\refe^1)$ as having negligible overlap with any cluster identified relative to $\refe^2$.  The overdensity $\mathcal{Y}^{\text{anom}}_{0}(\refe^1)$ is therefore induced by  the systematic $\refe^1\rightarrow\refe^2$, as can be qualitatively observed  from the overlap of the yellow contours and red contours in Fig.~\ref{fig:systematic} of \ref{sec:additional_lhc_systematic}. That is, we interpret $\mathcal{Y}^{\text{anom}}_{0}(\refe^1)$ as arising from a discrepancy in $\refe^1$'s background modelling   with respect to the underlying background distribution present in the test data $\test$.

\paragraph{Takeaway.}
\EagleEye\ provides a statistically principled way to quantify and localize discrepancies between two or more background model estimates. By identifying compact, multivariate overdensities in a test dataset that persist across heterogeneous background models, \EagleEye\ robustly detects anomalies even when the true background distribution is unknown. This capability complements existing anomaly-detection strategies in searches for new physics at the LHC, as well as in any experimental setting where density anomalies must be identified under background uncertainty.

An additional benefit of this approach would be the ability to construct multi-feature systematics maps that reveal how regions dominated by background-model differences respond to theoretical and modelling uncertainties. Such maps could identify and quantify areas of feature space that are most sensitive to theory model assumptions, thereby guiding targeted validation efforts, model refinement, and control-region studies.

\begin{figure*}[t]
\centering
\includegraphics[width=.8\textwidth]{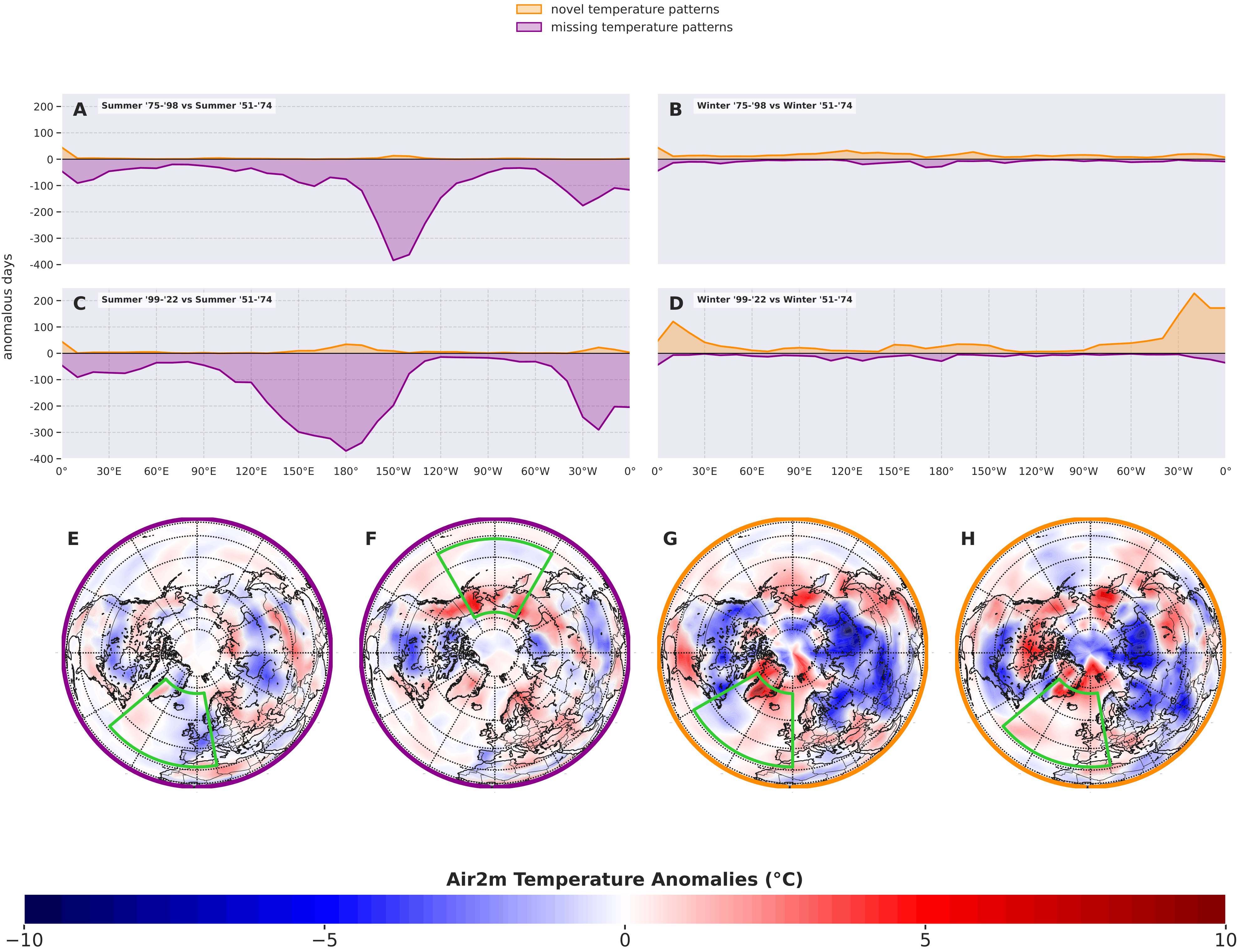} 
\caption{\textbf{Reanalysis temperature-pattern anomalies in JJA and DJF.}
\EagleEye\ applied to daily mean 2\,m air temperature anomaly fields (Air2m) from NCEP/NCAR Reanalysis~1 over the Northern Hemisphere, analysed separately for summer (JJA) and winter (DJF). Fields are de-trended and de-seasonalized per grid cell (Sec.~\ref{sec:Reanalisis1}) and analysed in moving longitudinal windows of width $60^\circ$ (green boxes), restricted to mid-latitudes $\phi\in[32^\circ\mathrm{N},72^\circ\mathrm{N}]$ following Springer \textit{et al.}~\cite{Springer2024}. Each day is mapped to a window-restricted anomaly-field vector, and nearest neighbours are computed using the same area-weighted, Gaussian-localized distance as in~\cite{Springer2024}. 
\textbf{(A-D)} Signed number of anomalous days per window for two comparisons against the early reference period (each period comprises 2130 seasonal days): 1975-1998 vs 1951-1974 (A,B) and 1999-2022 vs 1951-1974 (C,D). Positive values denote $\mathcal{Y}$-overdensities (emergent patterns; $\mathcal{X}\!\rightarrow\!\mathcal{Y}$ pass, IDE-pruned set $\IEo$), while negative values denote $\mathcal{X}$-overdensities (disappearing patterns; symmetric $\mathcal{Y}\!\rightarrow\!\mathcal{X}$ pass, IDE-pruned set $\IEu$). 
\textbf{(E-H)} Exemplar anomaly configurations for selected windows: average de-trended/de-seasonalized fields over the highest-scoring day in the corresponding window and its ten nearest neighbours {within the rep\^{e}chage anomaly set}. DJF exemplars are shown for windows centred at $30^\circ$W and $20^\circ$W (G,H), and JJA exemplars for windows centred at $180^\circ$ and $30^\circ$W (E,F).}
\label{fig:FIG02}
\end{figure*}
\subsection{Anomaly detection in time series: identifying localized  temperature anomalies in meteorology}
\label{sec:Reanalisis1}
In this section we demonstrate the use of \EagleEye on time-series data. 

\paragraph{Setup.}
We apply \EagleEye to daily mean 2\,m air temperature (Air2m) fields from the NCEP/NCAR Reanalysis~1 dataset~\cite{NCEP}, restricted to the Northern Hemisphere and analysed separately for winter (December-February; DJF) and summer (June-August; JJA). The gridded fields have resolution $2.5^\circ\times2.5^\circ$ (37 latitude bins $\times$ 144 longitude bins). Our goal is to detect \emph{changes in the recurrence} of spatial temperature patterns between earlier and later decades, while suppressing large-scale trends.

To emphasize changes in the \emph{recurrence} of spatial patterns we de-trend and de-seasonalize the Air2m field at each grid cell independently. For each season (winter and summer), we first subtract the seasonal mean of the corresponding year, thereby removing interannual shifts in the seasonal baseline. We then remove the residual seasonal cycle by subtracting, for each calendar day within the season, the across-year mean at that grid cell.

Following  Ref.~\cite{Springer2024}, we analyse the anomaly fields in a moving longitudinal window of width $60^\circ$ centered at longitude $\lambda_0$, i.e.\ $\lambda\in[\lambda_0-30^\circ,\lambda_0+30^\circ]$ with wrap-around at $0^\circ/360^\circ$. Within each window we restrict attention to Northern Hemisphere mid-latitudes, $\phi\in[32^\circ\mathrm{N},72^\circ\mathrm{N}]$, matching the spatial domain in~\cite{Springer2024}. For a window centered at $\lambda_0$, each day $t$ is represented by the vector of window-restricted temperature anomalies. \EagleEye\ anomaly scores $\boldsymbol{\Upsilon}_i$ are computed using the same distance as defined in~\cite{Springer2024}. Grid cells are area-weighted to account for the latitude dependence of grid-cell surface, and a Gaussian longitudinal kernel that gives more focus to the window centre; we refer to Ref.~\cite{Springer2024} for the explicit metric definition and parameter choices.

We partition the data set into three contiguous periods of equal length, each comprising  $2130$ seasonal days: a reference period (1951-1974) and two test periods (1975-1998 and 1999-2022).\footnote{We assign each winter season to   December-January-February (DJF), so ``DJF 1951'' includes December 1950; summer includes June, July and August (JJA). }
In all runs we use $K_M=100$ and an exceedance target $p_{\mathrm{ext}}=10^{-5}$, yielding  critical threshold $\PNval_+^* = 11.7$.

\paragraph{Workflow.}
For each window centre $\lambda$ and for each season separately, we run \EagleEye as a two-sample comparison between the reference set of daily window-restricted fields from 1951-1974 and the corresponding test set from either 1975-1998 or 1999-2022. The $\mathcal{X}\!\rightarrow\!\mathcal{Y}$ pass identifies \emph{emergent} patterns that are over-represented in the test period relative to the reference, while the symmetric $\mathcal{Y}\!\rightarrow\!\mathcal{X}$ pass identifies \emph{disappearing} patterns (equivalently, windows where the reference carries excess probability mass). For compactness, we summarize outcomes using the IDE-pruned set, which quantitatively represents the number of anomalous days per window. To display both directions in a single panel, we report signed counts, with positive values denoting $\mathcal{Y}$-overdensities (test over-represented; $\mathcal{X}\!\rightarrow\!\mathcal{Y}$ pass) and negative values denoting $\mathcal{X}$-overdensities (reference over-represented; $\mathcal{Y}\!\rightarrow\!\mathcal{X}$ pass). To assess the   effect of varying the placement of the reference interval, we repeat the analysis using reference sets obtained by shifting the start year between 1950 and 1951 while keeping the reference length fixed. The resulting variation in signed counts is negligible, so we report the mean outcome.

\paragraph{Outputs.}
Fig.~\ref{fig:FIG02} summarizes the moving-window results. \textbf{Panels A-D} report, for each longitude window and each period comparison, the signed number of anomalous days in the IDE-pruned sets (cardinality of $\IEo$ for $\mathcal{Y}$-overdensities, plotted as positive and cardinality of $\IEu$ for $\mathcal{X}$-overdensities, plotted as negative). \textbf{Panels E-H} provide qualitative exemplars of detected clusters: for selected window centres corresponding to prominent peaks in Panels A-D, we display the average de-trended and de-seasonalized anomaly field computed over (i) the day with the largest anomaly score in that window and (ii) its ten nearest neighbours within the corresponding rep\^{e}chage anomaly set (nearest neighbours evaluated in the same windowed feature space). Green boxes indicate the $60^\circ$ longitudinal windows used for the analysis.

\paragraph{Takeaways.}
Two robust results emerge. First, the sign structure differs by season: during summer (Panels A and C), negative values dominate, indicating \emph{disappearing} patterns-configurations common in 1951-1974 that become less frequent in later decades. During winter (Panels B and D), positive values dominate, indicating \emph{emergent} patterns that recur more often in the later periods than in the early reference. Second, the most recent period (1999-2022) exhibits stronger and more geographically localized shifts than 1975-1998, most notably over the Atlantic sector in winter, where a pronounced peak appears only in the later comparison. The exemplar composites (Panels E-H) confirm that these peaks correspond to coherent, window-localized temperature-anomaly configurations, i.e., unusually high temperatures ; the winter Atlantic exemplars emphasize strong positive anomalies within the window (with intense anomalies over Greenland), whereas the summer exemplars feature structured fields with both positive and negative anomalies.

\section*{Conclusion}
We introduced \EagleEye, a two-sample density-comparison method for detecting \emph{localized} differences between multivariate datasets. \EagleEye builds on a simple idea: under the null hypothesis that the two samples are drawn from the same distribution, the class labels of $k$ nearest neighbours behave approximately as Bernoulli trials. This yields a local test statistic whose maximisation over neighbourhood ranks provides sensitivity across anomaly scales. The resulting pipeline is conceptually transparent, deterministic, straightforwardly parallelisable, and computationally efficient, with performance governed primarily by a nearest-neighbour search, which can be made $O(n\log n)$ with tree-based standard approaches.

Across synthetic benchmarks, \EagleEye reliably localizes both excesses and deficits, remaining robust to changes in anomaly size, morphology, and sample imbalance. Beyond detection, the method produces \emph{interpretable} anomaly sets through iterative density equalization and multimodal rep\^{e}chage, and supports an estimate of irreducible background and signal purity within each detected mode.

We demonstrated the method on two real-world case studies. In a resonant-anomaly search using simulated LHC data, \EagleEye identifies evidence of a  new physical process which manifests as an over density that persists across multiple reference sets, enabling the separation of potential new-physics candidates from background mis-modelling. In climate analysis data, \EagleEye detects geographically and seasonally localized shifts in the recurrence of temperature-pattern configurations, highlighting both emergent and disappearing regimes beyond what is captured by global mean trends. Together, these applications show that \EagleEye’s strength lies not only in handling diverse data modalities, but also in its clear, deterministic, statistics-based formulation, which supports transparency and interpretability.

Several extensions to the \EagleEye method remain natural. Future work will be devoted to combining our approach with dimensionality reduction techniques and neural network–based feature learning/selection methods to identify the most informative feature space for anomaly detection (see, e.g., difFOCI \cite{pavasovic2025differentiablerankbasedobjectivebetter}).

Because \EagleEye requires only a multivariate point clouds (no specialised data format or model), it is broadly applicable wherever localized density discrepancies matter, with relevance far beyond scientific discovery, including potential applications in finance (fraud, regime change), healthcare (rare-event monitoring), cybersecurity (intrusion/drift detection), and industrial sensing (fault detection and predictive maintenance).

\section*{Code Availability}

The \EagleEye\ code, as well as the the code used to generate the results of this study are available at GitHub via the link: \url{https://github.com/sspring137/EagleEye}.

\newpage

\section*{Materials and Methods}
\setcounter{subsection}{0}
\renewcommand{\thesubsection}{M\arabic{subsection}}

\label{sec:methods}

\EagleEye\ is a deterministic, $k$NN-based two-sample procedure that localizes where two datasets differ in density. Given a \emph{reference} sample $\refe$ and a \emph{test} sample $\test$, the pipeline consists of three core steps: (i) \emph{flagging} points in $\test$ (and symmetrically in $\refe$) whose local neighborhood composition is inconsistent with the null hypothesis of equal underlying densities; (ii) \emph{iterative density equalization} (IDE), which prunes the flagged set to a collection of representative points by iteratively removing the strongest local discrepancies until the remaining samples become locally indistinguishable; and (iii) \emph{multimodal rep\^{e}chage}, which converts representatives into cluster-resolved anomaly sets intended for downstream inspection. As a post-processing step, we also describe an injection-based estimator of irreducible background and purity within each localized anomaly.

\subsection{Flagging of putative anomalous points}
\label{sec:flagging}

Let \(\refe = \{X_1, X_2, \dots, X_{n_\mathcal{X}}\}\) and \(\test = \{Y_1, Y_2, \dots, Y_{n_\mathcal{Y}}\}\) be two independent samples drawn from $d$-variate distributions. We denote the pooled set by \(\both = \refe \cup \test\).
For each point \(Y_i \in \test\), let \(\mathcal{N}_{\bar{k}}(Y_i) \subseteq \both\) be its \(\bar{k}\)-nearest neighbors under a chosen metric (e.g., Euclidean). Encode the neighbor memberships in the binary sequence
\[
\mathbf{b}_i \;=\; \bigl(b_i^1,\,b_i^2,\dots,b_i^{\bar{k}}\bigr),
\]
\[
b_i^k \;=\;
\begin{cases}
  0, & \text{if the \(k\)-th neighbor of \(Y_i\) lies in \(\refe\)}, \\
  1, & \text{if the \(k\)-th neighbor of \(Y_i\) lies in \(\test\)}.
\end{cases}
\]
Under the null hypothesis that \(\refe\) and \(\test\) are sampled from the same distribution, the baseline probability that a random neighbor belongs to $\test$ is
\begin{equation}
\phat \;=\; \frac{n_\mathcal{Y}}{n_\mathcal{X} + n_\mathcal{Y}}.
\label{eqn:phat}
\end{equation}
Define the cumulative test-neighbor count within rank $K$ as
\[
B(i,K) \;=\; \sum_{k=1}^{K} b_i^k.
\]
In the regime $K \ll n_\mathcal{X}+n_\mathcal{Y}$, we may approximate the distribution of  $B(i,K)$ under the null as
\[
B(i,K) \sim \mathrm{Binomial}(K,\phat).
\]
We note that the exact null is hypergeometric due to sampling without replacement. Details are discussed in \ref{sec:SI0}.

For each observed value \( B_{\text{obs}}(i,K) \), compute the right-tail $p$-value under the null and define the score
\[
\PNval(i,K) := -\log\Bigl( \mathrm{pval}\bigl( B_{\text{obs}}(i,K) \bigr) \Bigr).
\]
Finally, define the per-point anomaly score as the maximized statistic over neighborhood ranks up to $K_M$:
\begin{align}
\label{eq:NPval}
\PNval_i := \max_{1\le K \le K_M} \PNval(i,K).
\end{align}
The parameter $K_M$ controls the range of spatial scales probed: larger $K_M$ increases sensitivity to broader discrepancies at higher computational cost, whereas smaller $K_M$ restricts detection to more pronounced local imbalances between the two samples at lower cost. In practice we enforce
\[
K_{M} \;\le\; 0.05\, \min(n_\mathcal{X}, n_\mathcal{Y}),
\]
and for large datasets typically use $K_M=\mathcal{O}(10^2) / \mathcal{O}(10^3)$, since increasing $K_M$ beyond a few thousand yields diminishing returns \ref{sec:SI0b}.

We flag putative anomalous points in the test sample as
\begin{align}
\label{eq:OG}
\OG^+ := \left\{ Y_i \in \test \,\big|\, \PNval_i \geq \PNval^*_+ \right\},
\end{align}
where the critical threshold $\PNval^*_+$ is calibrated so that the \emph{per-point} null exceedance probability equals a user-chosen level $p_{\mathrm{ext}}$ (default $10^{-5}$). Because $\PNval_i$ is a maximum over $K\le K_M$, we calibrate $\PNval^*_+$ by Monte Carlo (Supplementary Information, \ref{sec:SI1c}). Intuitively, when $\phat=0.5$ this corresponds to identifying points whose neighborhood composition contains an anomalously large fraction of test points relative to a fair coin-flip baseline.

To localize relative deficits of $\test$ (equivalently, overdensities in $\refe$), we repeat the procedure with roles interchanged, $\test\leftrightarrow\refe$, yielding
\[
\refe^+ := \left\{ X_i \in \refe \,\big|\, \PNval_i \geq \PNval^*_- \right\}.
\]
When $n_\mathcal{X}=n_\mathcal{Y}$, symmetry implies $\PNval^*_+=\PNval^*_-$. An overdensity in $\refe$ corresponds to a local underdensity in $\test$ in the same region of feature space.

\subsection{Pruning via iterative density equalization (IDE)}
\label{sec:pruning}

The flagged sets $\OG^+$ and $\refe^+$ can include nearby non-anomalous points that inherit elevated scores because their neighborhoods partially intersect a true discrepancy region (a ``halo'' effect). IDE reduces this redundancy and sharpens localization by iteratively removing the strongest local discrepancy until the remaining data are locally consistent with the null.

We describe IDE for $\test$-side overdensities; the $\refe$-side procedure is obtained by swapping $\test\leftrightarrow\refe$. Initialize $\mathcal{Y}^{\mathrm{eq}}=\test$ and $\IEo=\emptyset$. Then iterate:
\begin{enumerate}
\item Compute $\PNval_i$ for all $Y_i\in\mathcal{Y}^{\mathrm{eq}}$ (with $\refe$ fixed).
\item Identify the point $Y_{i_{\max}}$ with the largest value of $\PNval$:  $i_{\max}=\argmax_{i\in \mathcal{Y}^\text{eq}} \PNval_i$.
\item Remove \(Y_{i_{\max}}\) and all its nearest neighbours (up to the next point belonging to \(\refe\)) from $\mathcal{Y}^\text{eq}$, adding them  to the set \(\IEo\). 
\end{enumerate}
Stop when $\max_{Y_i\in\mathcal{Y}^{\mathrm{eq}}}\PNval_i < \PNval^*_+$, at which point $\mathcal{Y}^{\mathrm{eq}}$ is locally density-equalized with respect to $\refe$ by construction, and $\IEo\subseteq\OG^+$ comprises representative overdensity points. Repeating symmetrically yields the equalized reference set $\refe^{\mathrm{eq}}$ and pruned representatives $\IEu\subseteq\refe^+$.

\subsection{Multimodal rep\^{e}chage}
\label{sec:repechage}

Multiple localized discrepancies may be present. We therefore first cluster the flagged sets (step~1) and then define, within each cluster, a robust score threshold informed by the IDE representatives.

We cluster $\OG^+$ (and analogously $\refe^+$) using Density Peaks Advanced (DPA)~\cite{DPA21} as implemented in \texttt{dadapy}~\cite{dadapy}. Let $\test^+_{\alpha}$ denote the subset of $\OG^+$ assigned to cluster label $\alpha$. For each cluster, define a cluster-specific threshold
\begin{equation}
\label{eq:quantile_repchage}
\PNval_\alpha^+ \;=\; \operatorname{quantile}\!\left( \{\PNval_i:\ Y_i \in \IEo \cap \test^+_{\alpha}\},\ q \right),
\end{equation}
with default $q=10^{-2}$. This guards against a single spuriously low representative score causing an overly permissive expansion of the anomaly set.

The rep\^{e}chage anomaly set for cluster $\alpha$ is then
\begin{align}
\test_\alpha^{\mathrm{anom}}
&:= \left\{ Y_i\in\test^+_\alpha\ \big|\ \PNval_i \ge \PNval_\alpha^+ \right\}.
\end{align}
The union $\bigcup_\alpha \test_\alpha^{\mathrm{anom}}\subset\test$ constitutes the final set of localized $\test$-overdensities. The analogous construction on $\refe$ yields $\bigcup_\alpha \refe_\alpha^{\mathrm{anom}}\subset\refe$, which identifies regions where the reference carries excess local mass (equivalently, where $\test$ exhibits a relative deficit).

\subsection{Estimating irreducible background and signal purity via injection}
\label{sec:injection}

After rep\^echage, the set $\bigcup_\alpha \test_\alpha^{\mathrm{anom}}$ will inevitably contain points that
are not anomalous, but that occur in the same feature space region of a density anomaly. For the case of $\test-$\textit{overdensities}, it is useful to consider that within the region of each anomaly $\alpha$ the test set is an admixture of two components: (i) points belonging to a density anomaly, which we call \emph{signal}, whose expected cardinality under the anomaly-generating process is $S_\alpha$; and (ii) points originating from the same (unknown) distribution as the reference set, which we call \emph{background points} (i.e., points that would be present even in the absence of an anomaly), whose expected cardinality is $B_\alpha$. In order to estimate the relative proportions of signal to background points in $ \test_\alpha^{\mathrm{anom}}$, we introduce the following \emph{injection scheme}: we inject individual samples from the reference set into the test set, thus creating a representative, traceable subset of non-anomalous points.  
The proportion of injected samples that is identified as anomalous by \EagleEye provides an estimate of the \emph{irreducible background}, i.e., the proportion of background points that are attributed to contributing to a density anomaly. 


In practice, we move each point $X_i \in \refe$ individually to $\test$, creating the temporary sets 
$\test \cup \{ X_i\}$ and $\refe \setminus \{ X_i\}$; we then recalculate the point's anomaly score and determine whether this value exceeds the threshold defined in Equation~\ref{eq:OG}, in which case the point is flagged as a potential anomaly. This happens when the injected point is within, or in close proximity to, an actual anomaly in $\test$.  This procedure yields the set:
\begin{align}
\label{eq:val}
\underline{\OG}^+ = \Big\{ X_i \in \refe \,\big|\, \PNval_i \geq \PNval^*_+ \;\text{, when;}\\ \test=\test\cup \{ X_i\} \text{ and } \refe=\refe\setminus\{ X_i\}\Big\}.
\end{align}
Naturally, the number of points in $\underline{\OG}^+$ increases if anomalies are in regions with higher background density.

For each anomaly $\alpha$,  analogously to the anomalous set $\test_\alpha^{\mathrm{anom}}$, we  define the \emph{injected set}:
\begin{align}
\test_\alpha^{\mathrm{inj}} := \Big\{ X_i \in \underline{\OG}^+ \,\Big|\, \PNval_i \geq \PNval^+_\alpha \Big\}.
\end{align}

The injected set $\test_\alpha^{\mathrm{inj}}$ can be used to obtain a quantitative measure of the \textit{signal-to-noise} ratio for an anomaly $\alpha$, as we show below in Equation~\ref{eqn:s/b}.  Its counterpart set, $\refe_\alpha^{\mathrm{inj}}$, represents underdense points in $\test$ corresponding to a local overdensity in $\refe$.

{
To quantify the presence of a signal relative to the background within an anomaly region $\alpha$, we consider the \emph{signal-to-noise ratio}
\(S_\alpha / \sqrt{B_\alpha},\)
where $S_\alpha$ denotes the expected number of signal points and $B_\alpha$ the expected number of background points contributing to $\test_\alpha^{\mathrm{anom}}$. Under a Poisson counting model for background points, $\sqrt{B_\alpha}$ is the natural fluctuation scale of the background, so $S_\alpha / \sqrt{B_\alpha}$ measures the excess signal in units of background variability. 
Large values indicate that the excess is substantial relative to background fluctuations, whereas values of order unity or smaller indicate that the excess is comparable to, or smaller than, typical background variability.}

We estimate this quantity using the following plug-in estimator (the derivation is
given in Section~\ref{sec:SI0c}):
\begin{align}
      \Lambda_\alpha := \widehat{\frac{S_\alpha}{\sqrt{B_\alpha}}} =  
    \frac{\card{\test_\alpha^{\mathrm{anom}}} - \left( \card{\test_\alpha^{\mathrm{inj}}}\frac{(n_\mathcal{Y} - \card{\IEo})}{(n_\mathcal{X} - \card{\IEu})}  \right) }{\sqrt{ \card{\test_\alpha^{\mathrm{inj}}}\frac{(n_\mathcal{Y} - \card{\IEo})}{(n_\mathcal{X} - \card{\IEu})}  } } \, .
    \label{eqn:s/b}
\end{align}
This estimator can be interpreted as follows: The signal, represented by the numerator, is the total number of points identified as anomalous, \(\card{\test_\alpha^{\mathrm{anom}}}\), minus the expected background contribution. The latter is estimated from the injection procedure, \(\card{\test_\alpha^{\mathrm{inj}}}\), as detailed above, scaled by an appropriate reweighting factor which accounts for the relative sizes of the background components in the test set and the reference set. 

{
In regions where the expected background count $B_\alpha$ is sufficiently large, the standardized excess  
$(\card{\test_\alpha^{\mathrm{anom}}} - B_\alpha)/\sqrt{B_\alpha}$
is approximately Normally distributed under a background-only model. In this regime, $\Lambda_\alpha$ may be interpreted as an approximate $z$-score and used as a heuristic measure of statistical significance \cite{Tanner2010Shorter,Cowan:2010js}.
Outside the large-background regime, $\Lambda_\alpha$ should be interpreted more generally as a standardized measure of discrepancy strength rather than a calibrated test statistic.}

A corresponding, counterpart for the derivation of a signal-to-noise with respect to $\refe-$\textit{overdensities} can be analogously obtained by the interchange of $(\test_\alpha^{\mathrm{anom}} ,\IEo , n_\test , \test_\alpha^{\mathrm{inj}}) \leftrightarrow (\refe_\alpha^{\mathrm{anom}}, \IEu, n_\refe , \refe_\alpha^{\mathrm{inj}})$ in Equation~\ref{eqn:s/b}. We additionally present an estimator for the \emph{signal-purity}  $S_\alpha/(S_\alpha+B_\alpha)$ ( Refs.~\cite{Fruhwirth:1990zas,cowan1998statistical}) in \ref{sec:SI0c}.

\textbf{}

\section*{Author contributions}
\printauthorcontributions

\section*{Acknowledgments}
We thank Abhishek Sharma for helpful discussions on collider aspects. SS and AL acknowledge financial support from Regione Friuli Venezia Giulia (project F53C22001770002); AL also acknowledges funding from the Italian National Centre for HPC (Grant No. CN00000013). AS was partially supported by “DS4ASTRO: Data Science methods for Multi-Messenger Astrophysics $\&$ Multi-Survey Cosmology” under the PRO3 ‘Programma Congiunto’ (DM n. 289/2021) of the Italian Ministry for University and Research. AS and RT acknowledge funding from Next Generation EU under the National Recovery and Resilience Plan, Investment PE1 – Project FAIR “Future Artificial Intelligence Research”; this resource was co-financed by the Next Generation EU [DM 1555 del 11.10.22]. RT is partially supported by Fondazione ICSC, Spoke 3 “Astrophysics and Cosmos Observations”, Piano Nazionale di Ripresa e Resilienza Project ID CN00000013 “Italian Research Center on High-Performance Computing, Big Data and Quantum Computing”, funded by MUR Missione 4 Componente 2 Investimento 1.4: Potenziamento strutture di ricerca e creazione di “campioni nazionali di R$\&$S (M4C2-19 )” - Next Generation EU (NGEU). GC is supported by the European Union’s Horizon Europe research and innovation program under the Marie Sklodowska-Curie COFUND Postdoctoral Programme grant agreement No. 101081355- SMASH and by the Republic of Slovenia and the European Union from the European Regional Development Fund. Disclaimer: Co-funded by the European Union; views and opinions expressed are however those of the author(s) only and do not necessarily reflect those of the European Union or European Research Executive Agency. Neither the European Union nor the granting authority can be held responsible for them. MA acknowledges partial support from the UK Engineering and Physical Sciences Research Council [EP/W015080/1, EP/W522673/1] and from the European Union’s Horizon 2020 research and innovation programme under European Research Council Grant Agreement No 101002652 (PI K. Mandel) and Marie Skłodowska-Curie Grant Agreement No 873089. HH was supported by the Research Council of Finland Flagship of Advanced Mathematics for Sensing Imaging and Modelling grant 359183.

\clearpage

\bibliographystyle{unsrtnat}
\bibliography{cleaned_references_final.bib}

\clearpage


\setcounter{figure}{0}
\renewcommand{\thefigure}{S\arabic{figure}}

\renewcommand{\thesection}{\textit{SI Appendix S}\arabic{section}}
\renewcommand{\thesubsection}{\thesection\alph{subsection}}
\setcounter{section}{-1}
\setcounter{subsection}{0}
\newcommand{\SIsection}[1]{%
  \section{#1}%
}

\newcommand{\SIsubsection}[1]{%
  \subsection{#1}%
}

\makeatletter
\renewcommand{\@seccntformat}[1]{%
  \csname the#1\endcsname\quad}
\makeatother

\setcounter{table}{0}
\renewcommand{\thetable}{S\arabic{table}}

\setcounter{equation}{0}
\renewcommand{\theequation}{ES\arabic{equation}}

\clearpage
\onecolumn
\begin{center}
\Large \textbf{Supporting information (SI) for "Detecting Localized Density Anomalies in Multivariate Data via Coin-Flip Statistics"}
\end{center}
\vspace{1em}

\SIsection{Methodology: Technical additions}
\label{sec:SI0}
This section provides technical additions that complement the methodology in the main text. SI0 is organised into four parts: \ref{sec:SI0a} discusses the expected maximum number of Binomial successes from $N$ trials  which is later used to investigate the properties of our critical threshold;  \ref{sec:SI0b} motivates our choice of the maximum neighbourhood rank $K_M$ and the Binomial approximation; \ref{sec:SI1c}  describes in detail the derivation of the threshold $\Upsilon^*_+$; and \ref{sec:SI0c} derives the injection-based estimators used to quantify signal-to-noise and related summary statistics for each detected anomaly mode.

\SIsubsection{Expected maximum number of Binomial successes}
\label{sec:SI0a}
Consider \(N\) independent and identically distributed random variables
\(\{X_i\}_{i=1}^N\), each following a \(\mathrm{Binomial}(n,p)\) distribution.
Let the cumulative distribution function (CDF) of each \(X_i\) be
\[
F_X(m) \;=\; \Pr(X_i \le m).
\]
Then the CDF of the maximum
\[
X^\dag \;=\;\max_{1\le i\le N} X_i
\]
is obtained by noticing that the random variables $X_i$ are independent:
\[
\begin{aligned}
F_{X^\dag}(m)
&= \Pr\bigl(X^\dag \le m\bigr)
= \Pr\bigl(X_1 \le m,\;X_2 \le m,\;\dots,\;X_N \le m\bigr) \\[0.3em]
&= \prod_{i=1}^N \Pr(X_i \le m)
= \bigl[F_X(m)\bigr]^N.
\end{aligned}
\]

To compute the expectation of a nonnegative integer--valued random variable \(Y\),
we use the tail-sum identity:
\[
E[Y]
= \sum_{y=0}^\infty y\,\Pr(Y=y)
= \sum_{k=1}^\infty \sum_{y=k}^\infty P(Y=y)
= \sum_{k=1}^\infty \Pr(Y \ge k)
= \sum_{m=0}^\infty \bigl[1 - \Pr(Y \le m)\bigr].
\]
Applying this to the maximum \(Y=X^\dag\), which takes values in \(\{0,1,\dots,n\}\), gives
\[
\begin{aligned}
E[X^\dag]
&= \sum_{m=0}^{n-1} \Pr(X^\dag > m)
= \sum_{m=0}^{n-1} \bigl[1 - F_{X^\dag}(m)\bigr] \\[0.3em]
&= \sum_{m=0}^{n-1} \Bigl[\,1 - \bigl(F_X(m)\bigr)^N\Bigr].
\end{aligned}
\]
See Ref.~\cite{david2003order} for further details.

\SIsubsection{Selection of maximum neighbourhood rank $K_M$}
\label{sec:SI0b}
We first notice that the model of the test statistics described in \ref{sec:methods} is a very accurate approximation, but is not exact.
While the test statistic \( B(i,K) \) is modelled as a Binomial distribution under the null hypothesis, the actual process follows a hypergeometric distribution because the nearest neighbours are selected without replacement from the combined set $\mathcal{U}$.
The Binomial distribution serves as an excellent approximation when the total number of points \( n_1 + n_2 \) is large relative to \( K \), ensuring that the difference between the two distributions becomes negligible in practice \citep{brunk1968teacher}.

\begin{figure}[h!]
    \centering
    \includegraphics[width=\linewidth]{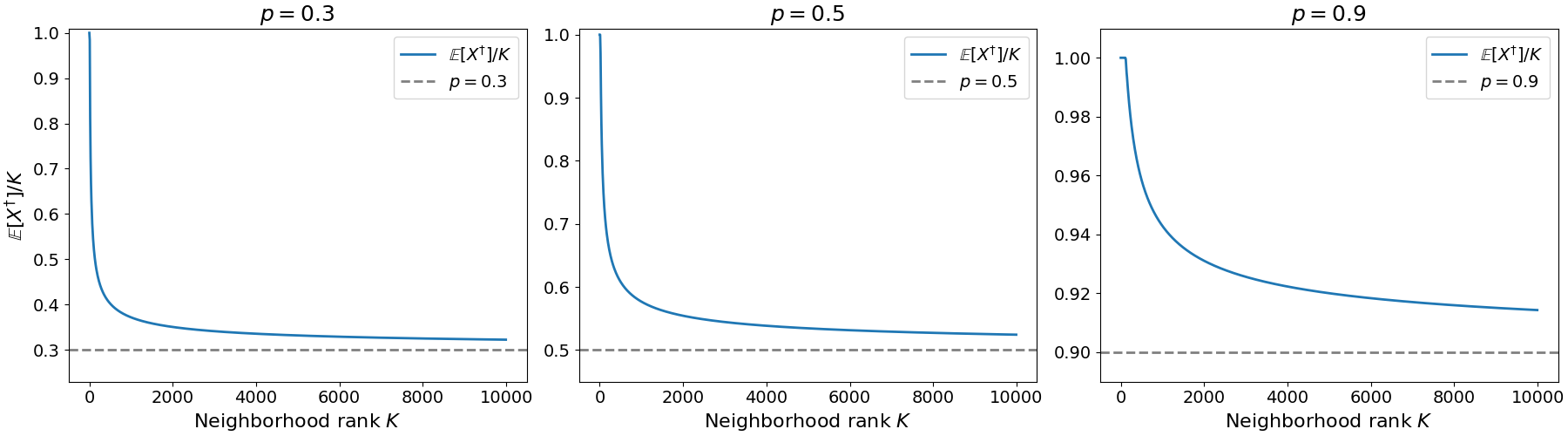}
    \caption{\textbf{Ratio of the expected maximum number of Binomial successes $X^\dag$ to the neighbourhood rank $K$}, computed from $5\times10^5$ samples, as a function of $K$ for three different success probabilities: (a) $p=0.3$, (b) $p=0.5$, and (c) $p=0.9$. Dashed black lines indicate the corresponding baseline probability $p$. As $K$ increases, $\mathbb{E}[X^\dag]/K$ decays monotonically and converges toward $p$. }
    \label{fig:F_binom}
\end{figure}

A key consideration when choosing the maximum neighbourhood rank $K_M$ is the behaviour of the detection performance as a function of $K_M$. Fig.~\ref{fig:F_binom} shows $\mathbb{E}[X^\dag]/K$ as a function of $K$. As $K$ grows, this critical fraction decays monotonically and asymptotically approaches the underlying success probability $p$. The ratio decreases very quickly for small $K_M$, but then decreases only slowly, implying that setting $K_M$ above a few thousand yields only marginal improvement. This may be useful for detecting very extended anomalies, although such broad discrepancies are often also captured by standard global shift/anomaly-detection methods. Therefore, for the purpose of this work, which is detecting \emph{localized} anomalies, we set \( K_M \) to $5\%$ of the data size for small datasets, and to a few thousand for large datasets, balancing detection power against computational cost.

\begin{figure}
    \centering
    \includegraphics[width=\linewidth]{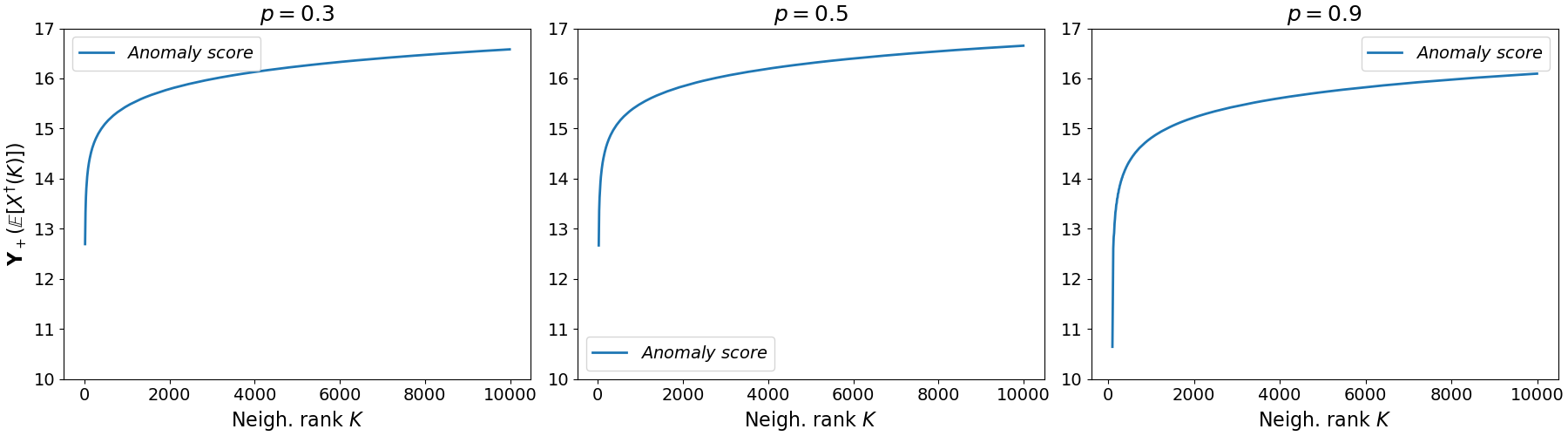}
    \caption{\textbf{Evolution of the anomaly score $\PNval_+$ for the expected maximum Binomial count}, computed by substituting $B_{\mathrm{obs}}(i,K)$ in \ref{eq:NPval} with $\mathbb{E}[X^\dag(K)]$ for three different success probabilities: (a) $p=0.3$, (b) $p=0.5$, and (c) $p=0.9$.}
    \label{fig:F_crit} 
\end{figure}

Finally, Fig.~\ref{fig:F_crit} illustrates how the anomaly score $\PNval_+$ of the expected maximum Binomial count, computed by substituting $B_{\mathrm{obs}}(i,K)$ in \ref{eq:NPval} with $\mathbb{E}[X^\dag(K)]$, evolves as a function of $K_M$. Consistent with Fig.~\ref{fig:F_binom}, the anomaly score increases only slowly for large $K_M$, reinforcing our choice to cap $K_M$ at $\mathcal{O}(10^{2})$ / $\mathcal{O}(10^{3})$  to balance sensitivity against computational efficiency. Moreover, the shallow increase of $\PNval_+(\mathbb{E}[X^\dag(K)])$ with $K_M$ suggests that detection performance remains robust to moderate variations in hyperparameter settings.

\SIsubsection{Null-threshold setting for the anomaly score}
\label{sec:SI1c}

EagleEye flags putative anomalies using the pointwise anomaly score
\[
\Upsilon_i \;=\;\max_{1\le K\le K_M}\Upsilon(i,K),
\qquad
\Upsilon(i,K)\;:=\;-\log\!\Big(\mathrm{pval}\big(B_{\mathrm{obs}}(i,K)\big)\Big),
\]
where $B_{\mathrm{obs}}(i,K)=\sum_{k=1}^{K} b_i^k$ is the observed number of test-sample neighbors among the first $K$ nearest neighbors of point $i$ (Materials and Methods). Under the null hypothesis that $\mathcal{X}$ and $\mathcal{Y}$ are drawn from the same distribution, each neighbor-indicator $b_i^k$ is modeled as a Bernoulli random variable with success probability
\[
\hat p=\frac{n_\mathcal{Y}}{n_\mathcal{X}+n_\mathcal{Y}},
\]
so that $B(i,K)\sim\mathrm{Binomial}(K,\hat p)$ is an accurate approximation in the large-sample regime where $K\ll n_\mathcal{X}+n_\mathcal{Y}$ (see \ref{sec:SI0b} above) for the binomial vs hypergeometric discussion and the rationale for $K_M$). For overdensity detection in the $\mathcal{X}\!\rightarrow\!\mathcal{Y}$ pass, we use the right-tail p-value
\[
\mathrm{pval}\big(B_{\mathrm{obs}}(i,K)\big)\;=\;\Pr\!\left[B(i,K)\ge B_{\mathrm{obs}}(i,K)\right].
\]
For deficit detection, we perform the symmetric pass $\mathcal{Y}\!\rightarrow\!\mathcal{X}$, which exchanges the roles of the samples and uses the same calibration logic.

We choose the critical score threshold $\Upsilon^{*}_{+}$ by controlling its \emph{per-point} exceedance probability under the null:
\[
\Pr_{\mathrm{null}}\!\left(\Upsilon_i \ge \Upsilon^{*}_{+}\right)=p_{\mathrm{ext}}.
\]
Thus, when no density discrepancy is present, only a fraction $p_{\mathrm{ext}}$ of points are expected to be flagged in the initial stage. The default value used in this work is $p_{\mathrm{ext}}=10^{-5}$.

Because $\Upsilon_i$ is defined as a maximum over $K\le K_M$, its null distribution is not the tail of a single Binomial variate. We therefore calibrate $\Upsilon^{*}_{+}$ by Monte Carlo using Bernoulli vectors that preserve the dependence across $K$:
\begin{enumerate}
\item Fix $(K_M,\hat p,N_{\mathrm{bern}})$.
\item For $i=1,\dots,N_{\mathrm{bern}}$, draw an i.i.d.\ Bernoulli vector
$\mathbf{b}_i=(b_i^1,\ldots,b_i^{K_M})$ with $b_i^k\sim \mathrm{Bernoulli}(\hat p)$.
Using the cumulative counts $B_i(K)=\sum_{k=1}^{K} b_i^k$, compute the anomaly score $\Upsilon_i$.
\item Set $\Upsilon^{*}_{+}$ to the empirical $(1-p_{\mathrm{ext}})$ quantile of $\{\Upsilon_i\}_{i=1}^{N_{\mathrm{bern}}}$.
\end{enumerate}

\SIsubsection{Derivation of the signal-to-background ratio estimator}
\label{sec:SI0c}

Following from the definitions of signal and background: 
\begin{align*}
S_\alpha+B_\alpha&=\left|\test_\alpha^{\text {anom }}\right| \nonumber\\
\\
\Rightarrow \frac{S_\alpha}{S_\alpha+B_\alpha}&=\frac{\left|\test_\alpha^{\text {anom }}\right|-B_\alpha}{\left|\test_\alpha^{\text {anom }}\right|} \nonumber
\end{align*}
The number of background samples in the anomalous test region $\alpha, B_\alpha$, is of course not known. However, we know that the test background $B_y$ and the reference background $B_x$ share the same distribution. $B_y$ is the non-anomalous test component, and $B_x$ is the non-anomalous component of the reference set. Via the injection scheme outlined in Materials and Methods ~\ref{sec:injection} we obtain the following (in expectation):
$$
\frac{B_\alpha}{B_y}=\frac{\left|\test_\alpha^{\mathrm{inj}}\right|}{B_x}
$$
and thus
$$
B_\alpha=\frac{\left|\test_\alpha^{\mathrm{inj}}\right|}{B_x} B_y.
$$
We can estimate the size of the test background component $B_y$ by subtracting the estimated size of the test overdensities from the total test size, i.e.
$$
B_y \approx n_\mathcal{Y}-\left|\IEo\right|.
$$
The size of the reference background component $B_x$ can be estimated by subtracting the estimated size of the reference overdensities from the total reference size, i.e.,
$$
B_x \approx n_\mathcal{X}-\left|\IEu\right|.
$$
Leading to the estimation:
$$
\frac{\widehat{S_\alpha}}{\sqrt{B_\alpha}}=\frac{\left|\mathcal{Y}_\alpha^{\text {anom }}\right|-\left(\left|\mathcal{Y}_\alpha^{\text {inj }}\right| \frac{\left(n_{\mathcal{Y}}-|\hat{\mathcal{Y}}+|\right)}{\left(n_{\mathcal{X}}-|\hat{\mathcal{X}}+|\right)}\right)}{\sqrt{\left|\mathcal{Y}_\alpha^{\text {inj }}\right| \frac{\left(n_{\mathcal{Y}}-|\hat{\mathcal{Y}}+|\right)}{\left(n_{\mathcal{X}}-|\hat{\mathcal{X}}+|\right)}}}
$$
An estimator of \emph{signal-purity} also directly follows: 
\begin{align}
    \widehat{\frac{S_\alpha}{S_\alpha+B_\alpha}} &= 
    \frac{\card{\test_\alpha^{\mathrm{anom}}} - \left( \card{\test_\alpha^{\mathrm{inj}}}\frac{(n_\mathcal{Y} - \card{\IEo})}{(n_\mathcal{X} - \card{\IEu})}  \right) }{{\card{\test_\alpha^{\mathrm {anom}}}}} \, .
    \label{eq:s/s+b}
\end{align}
If the anomaly is found in a region of low background, then $S_\alpha/(S_\alpha+B_\alpha) \sim 1$, while a value $S_\alpha/(S_\alpha+B_\alpha) \ll 1$  indicates that the anomaly is immersed in a region of large background.

\SIsection{Additional details for the synthetic result presented in Results \ref{sec:synthetic_uniform_gaussians}}
\label{sec:synth_data_details}
In this section we expand upon the synthetic walk though reported in the main text Results \ref{sec:synthetic_uniform_gaussians}.  
This example is intentionally a simple, controlled end-to-end walkthrough (uniform background with well-separated localized components). Hyperparameter sensitivity and more challenging synthetic settings are reported separately in \ref{sec:SI1} and \ref{sec:SI1g_manifold}. The scripts, configuration files, and random seed used to generate the walkthrough figures are available in the public repository cited in the main text (\url{https://github.com/sspring137/EagleEye}).
\SIsubsection{Data set generation}
We generate two three-dimensional samples, a reference set $\mathcal{X}$ and a test set $\mathcal{Y}$, each with $N_{\mathcal{X}}=N_{\mathcal{Y}}=5\times10^{4}$ points. Each sample is constructed as a mixture of a uniform background and a small number of isotropic Gaussian components. In this synthetic \emph{walkthrough} example we refer to the points drawn from the injected Gaussian components as \emph{signal points} (known by construction) and use them for recovery accounting. Background points are drawn i.i.d.\ from a uniform distribution over the bounded domain
\[
x \sim \mathrm{Unif}(\Omega), \qquad \Omega=[-100,100]^3,
\]
with the same $\Omega$ used for both $\mathcal{X}$ and $\mathcal{Y}$. We inject isotropic Gaussian components into each sample. For the test sample $\mathcal{Y}$, we draw $n^{(Y)}_\alpha$ points from each of seven Gaussians
\[
y \sim \mathcal{N}\!\left(\mu^{(Y)}_\alpha,\sigma^{2(Y)}_\alpha I_3\right), \qquad \alpha=1,\dots,7,
\]
and for the reference sample $\mathcal{X}$, we draw $n^{(X)}_\beta$ points from each of three Gaussians
\[
x \sim \mathcal{N}\!\left(\mu^{(X)}_\beta,\sigma^{2(X)}_\beta I_3\right), \qquad \beta=1,\dots,3.
\]
All remaining points are drawn uniformly from $\Omega$, so that each sample contains exactly $50{,}000$ points after injection. Points drawn from the Gaussian components are \emph{signal points} (known by construction) and are used to quantify recovery in Table~\ref{tab:S1}. The component sizes are:
\[
n^{(Y)}_\alpha \in \{50,100,200,300,500,700,900\}, \qquad 
n^{(X)}_\beta \in \{100,300,700\},
\]
so that $n_\mathcal{Y}^{\mathrm{uni}}=47{,}250$ and $n_\mathcal{X}^{\mathrm{uni}}=48{,}900$ uniform background points.

The Gaussian centres are fixed and well separated in $\mathbb{R}^3$. For completeness, the centres are:
\[
\begin{aligned}
\{\mu^{(Y)}_\alpha\}_{\alpha=1}^7 &= \{
(-7.5,-7.5,-7.5),\ (7.5,7.5,7.5),\ (65,0,0),\ (0,65,0),\ (0,0,65),\ (-80,0,0),\ (0,-80,0)\},\\
\{\mu^{(X)}_\beta\}_{\beta=1}^3 &= \{
(-22.5,-22.5,-22.5),\ (0,0,-80),\ (95,0,0)\}.
\end{aligned}
\]
The standard deviations for the $\mathcal{Y}$-components are assigned in increasing order with component size:
\[
(50,100,200,300,500,700,900)\ \longmapsto\ \sigma^{(Y)} \in (1,2,3,4,5,6,7),
\]
so that $\sigma^{2(Y)}_\alpha \in \{1^2,\ldots,7^2\}$. For the three $\mathcal{X}$-components (sizes $(100,300,700)$, in this order), we use standard deviations
\[
\sigma^{(X)} \in (3,2,1),
\]
so that $\sigma^{2(X)}_\beta \in \{3^2,2^2,1^2\}$.

The seven Gaussian components injected into $\mathcal{Y}$ define \emph{$\mathcal{Y}$-overdensity} regions. The three Gaussian components injected into $\mathcal{X}$ define \emph{$\mathcal{X}$-overdensity} regions; when \EagleEye\ is run with $\mathcal{X}$ as reference and $\mathcal{Y}$ as test, these appear as \emph{relative deficits} of $\mathcal{Y}$ with respect to $\mathcal{X}$ (since $\mathcal{X}$ carries extra probability mass there while $\mathcal{Y}$ does not). In the synthetic walkthrough, overdensities and deficits are obtained via the two symmetric passes $\mathcal{X}\!\rightarrow\!\mathcal{Y}$ and $\mathcal{Y}\!\rightarrow\!\mathcal{X}$ (main text).

\SIsubsection{\EagleEye hyperparameter setup}
We report \EagleEye-specific hyperparameters. Namely $n_\mathcal{X}=n_\mathcal{Y}=50{,}000$, hence $\hat p=0.5$, and we set $K_M=500$. Using $N_\text{bern}=10^{6}$ Bernoulli sequences and $p_{\mathrm{ext}}=10^{-5}$ yields
\[
\Upsilon^{*}_{+}=14.04.
\]
Because $n_\mathcal{X}=n_\mathcal{Y}$, the same threshold applies to the symmetric $\mathcal{Y}\!\rightarrow\!\mathcal{X}$ pass in this example.

\SIsubsection{Extended figure: Full synthetic walkthrough (Panels A--F)}
\label{sec:SI1d}
\begin{SCfigure*}[\sidecaptionrelwidth][t!]
\centering
\includegraphics[width=.8\textwidth]{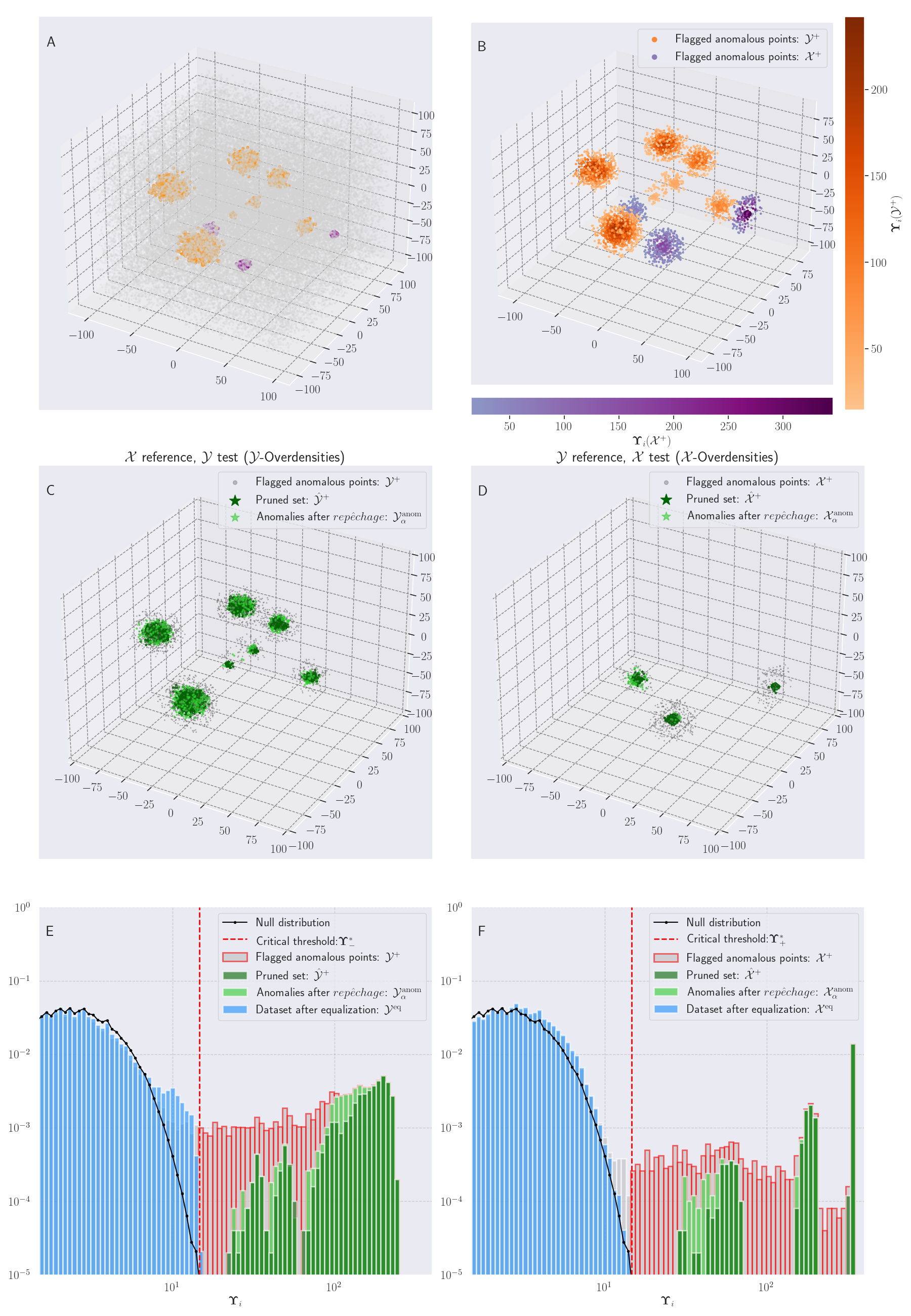} 
    \caption{\EagleEye\ detection of localized density anomalies in a uniform background.
\textbf{(A)} Constructed samples in feature space: injected $\mathcal{Y}$-overdensities and regions that appear as relative deficits of $\mathcal{Y}$ (obtained via the symmetric pass).
\textbf{(B)} Flagged sets ($\mathcal{Y}^{+}$ in the $\mathcal{X}\!\rightarrow\!\mathcal{Y}$ pass; $\mathcal{X}^{+}$ in the $\mathcal{Y}\!\rightarrow\!\mathcal{X}$ pass).
\textbf{(C--D)} Local anomalies after Iterative Density Equalization (IDE; dark green; $\IEo$ or $\IEu$) and rep\^{e}chage (light green; $\cup_{\alpha} Y^{\alpha}_{\mathrm{anom}}$ or $\cup_{\alpha} X^{\alpha}_{\mathrm{anom}}$).
\textbf{(C)} $\mathcal{X}\!\rightarrow\!\mathcal{Y}$ pass: $\mathcal{Y}$-overdensities.
\textbf{(D)} $\mathcal{Y}\!\rightarrow\!\mathcal{X}$ pass: $\mathcal{X}$-overdensities, which appear as relative deficits of $\mathcal{Y}$ with respect to $\mathcal{X}$.
\textbf{(E--F)} Distributions of the anomaly score ($\PNval_i$). Black: Monte Carlo null (Bernoulli sequences; \ref{sec:SI1c}). Grey: full sample. Dark green: IDE-pruned set. Light green: rep\^{e}chage set. Blue: equalized remainder, consistent with the null.}
    \label{fig:supp_SI1}
\end{SCfigure*}

 Fig.~\ref{fig:supp_SI1} provides an expanded visualization of the synthetic walk through, illustrating (i) how localized density discrepancies are identified in feature space and (ii) how the anomaly-score distribution evolves across the successive stages of the pipeline. Overdensities in the test sample are obtained in the $\mathcal{X}\!\rightarrow\!\mathcal{Y}$ pass; relative deficits of the test sample (corresponding to overdensities in the reference sample) are obtained via the symmetric $\mathcal{Y}\!\rightarrow\!\mathcal{X}$ pass. Below we detail each panel with a comprehensive discussion:

\begin{itemize}
\item \textbf{Panel A (construction).} The reference set $\mathcal{X}$ and test set $\mathcal{Y}$ are shown in feature space, comprising a uniform background on $\Omega=[-100,100]^3$ plus injected isotropic Gaussian components.
\item \textbf{Panel B (flagging).} Points are flagged when $\Upsilon_i\ge \Upsilon^{*}_{+}$, where $\Upsilon^{*}_{+}$ is calibrated under the null at exceedance probability $p_{\mathrm{ext}}$. This yields the flagged sets $\mathcal{Y}^{+}$ in the $\mathcal{X}\!\rightarrow\!\mathcal{Y}$ pass and $\mathcal{X}^{+}$ in the symmetric $\mathcal{Y}\!\rightarrow\!\mathcal{X}$ pass. At this stage, points in the local vicinity of an injected component may also be flagged due to neighborhood mixing (``halo'' points).
\item \textbf{Panel C (IDE + rep\^{e}chage: $\mathcal{X}\!\rightarrow\!\mathcal{Y}$).} Iterative Density Equalization (IDE) prunes the flagged set to a representative subset $\IEo$ (dark green), suppressing halo points while retaining points that best capture the local density discrepancy. The final rep\^{e}chage anomaly sets $\{\mathcal{Y}_{\mathrm{anom}}^{\alpha}\}$ (light green) are then obtained by clustering the flagged points and, within each cluster, selecting points co-located with the local anomaly cluster using the quantile rule with parameter $q$ (here $q=10^{-2}$). Panel C visualizes these outputs for $\mathcal{Y}$-overdensities detected in the $\mathcal{X}\!\rightarrow\!\mathcal{Y}$ pass.
\item \textbf{Panel D (IDE + rep\^{e}chage: $\mathcal{Y}\!\rightarrow\!\mathcal{X}$).} The same IDE and rep\^{e}chage procedure is applied in the symmetric pass, yielding $\IEu$ and $\{\mathcal{X}_{\mathrm{anom}}^{\alpha}\}$. Panel D visualizes these outputs for $\mathcal{X}$-overdensities, which appear as relative deficits of $\mathcal{Y}$ with respect to $\mathcal{X}$.
\item \textbf{Panels} \textbf{E}–\textbf{F} compare anomaly-score distributions $\Upsilon_i$ across successive stages of the pipeline to a calibrated null (black curve), generated from Monte Carlo Bernoulli sequences of length $K_M$ with success probability $\hat p$ using the same $(K_M,\hat p)$ as in the analysis. Grey histograms show scores for the full sample under test (all $\mathcal{Y}$ in panel E; all $\mathcal{X}$ in panel F), with flagged points highlighted ($\mathcal{Y}^{+}$ or $\mathcal{X}^{+}$). Dark green shows the IDE-pruned representatives ($\IEo$ or $\IEu$), and light green shows the union of cluster-resolved repechage anomaly sets ($\bigcup_{\alpha}\mathcal{Y}_{\mathrm{anom}}^{\alpha}$ or $\bigcup_{\alpha}\mathcal{X}_{\mathrm{anom}}^{\alpha}$). Blue shows the post-IDE equalized subset ($Y{\mathrm{eq}}$ or $X_{\mathrm{eq}}$), which closely follows the null overlay and serves as an internal diagnostic that detectable density discrepancies have been removed.
\end{itemize}
Supplementary Table~\ref{tab:S1} provides a component-by-component accounting of how points propagate through the \EagleEye\ pipeline in the synthetic walkthrough. For each detected local mode, we report the number of points retained after (i) \emph{flagging} ($\mathcal{Y}^{+}$ or $\mathcal{X}^{+}$), (ii) \emph{IDE pruning} (representatives $\IEo$ or $\IEu$), and (iii) \emph{rep\^{e}chage} (cluster-resolved anomaly sets $Y^{\alpha}_{\mathrm{anom}}$ or $X^{\alpha}_{\mathrm{anom}}$). Since the data are simulated, membership in an injected Gaussian component is known by construction; in Table~\ref{tab:S1}, the number of retained \emph{signal points} is reported in parentheses at each stage. Flagging is intentionally inclusive: it aims to capture all regions where the test sample departs from the reference, and can therefore include nearby background points (``halo'' points). IDE then removes redundant and peripherally flagged points, yielding a set of representatives that localize each discrepancy in feature space. Finally, rep\^{e}chage\ expands each representative cluster into a resolved anomaly set intended for downstream inspection. Table~\ref{tab:S1} also reports an estimate of the \emph{irreducible background} retrieved within each rep\^{e}chage set. This is obtained via the injection-based procedure described in \ref{sec:SI0c} and as specified in Eq.~\ref{eq:s/s+b}.  For reference, Table~\ref{tab:S1} also reports in parentheses the \emph{empirical purity} computed directly from the known signal membership.
\begin{table}[ht]
\centering
\resizebox{\textwidth}{!}{
\begin{tabular}{lcccccccccc}
\toprule
 & $\mathcal{Y}_{\alpha=0}$ & $\mathcal{Y}_{\alpha=1}$ & $\mathcal{Y}_{\alpha=2}$ & $\mathcal{Y}_{\alpha=3}$ & $\mathcal{Y}_{\alpha=4}$ & $\mathcal{Y}_{\alpha=5}$ & $\mathcal{Y}_{\alpha=6}$ & $\mathcal{X}_{\alpha=0}$ & $\mathcal{X}_{\alpha=1}$ & $\mathcal{X}_{\alpha=2}$ \\
\midrule
\textbf{Signal} & 50 (50) & 100 (100) & 200 (200) & 300 (300) & 500 (500) & 700 (700) & 900 (900) & 100 (100) & 300 (300) & 700 (700) \\
\textbf{Flagged} & 91 (50) & 167 (100) & 326 (200) & 507 (300) & 770 (500) & 1015 (699) & 1320 (897) & 227 (100) & 608 (300) & 862 (700) \\
\textbf{IDE-pruned} & 52 (50) & 69 (68) & 153 (149) & 230 (214) & 359 (338) & 526 (490) & 669 (588) & 89 (72) & 295 (290) & 702 (699) \\
\textbf{Rep\^{e}chage} & 62 (50) & 93 (92) & 211 (194) & 308 (281) & 537 (472) & 688 (610) & 1072 (861) & 146 (99) & 309 (300) & 704 (700) \\
\textbf{Irr.\ Bkg.\ est.}  & 11 & 6 & 19 & 25 & 67 & 85 & 211 & 22 & 10 & 5 \\
\textbf{Purity}  & 0.83 (0.81) & 0.94 (0.99) & 0.91 (0.92) & 0.92 (0.91) & 0.88 (0.88) & 0.88 (0.89) & 0.81 (0.80) & 0.85 (0.68) & 0.97 (0.97) & 0.99 (0.99) \\
\bottomrule
\end{tabular}
}
\caption{Support table for the synthetic walkthrough in Fig.~\ref{fig:supp_SI1}. The first seven columns correspond to injected $\mathcal{Y}$-overdensities (detected in the $\mathcal{X}\!\rightarrow\!\mathcal{Y}$ pass), while the last three columns correspond to injected $\mathcal{X}$-overdensities (which appear as relative deficits of $\mathcal{Y}$ and are detected via the symmetric $\mathcal{Y}\!\rightarrow\!\mathcal{X}$ pass). For each component, entries report the number of retained points at each stage; values in parentheses give the number of retained {signal points} (known by construction). The penultimate row reports the injection-based estimate of irreducible background. The final row reports the corresponding purity estimate from Eq.~\ref{eq:s/s+b}, with the  computed from the known signal membership shown in parentheses.}
\label{tab:S1}
\end{table}

Overall, Table~\ref{tab:S1} highlights three complementary aspects of the method: near-complete capture of discrepant regions at the flagging stage, parsimonious localisation via IDE representatives, and compact, mode-resolved anomaly sets after rep\^{e}chage\ with high estimated purity for the larger injected components.

\SIsection{Sensitivity studies}
\label{sec:SI1}
In this section, we systematically benchmark the robustness of the \EagleEye\ pipeline beyond the simple walkthrough setting of Results~\ref{sec:synthetic_uniform_gaussians} and \ref{sec:synth_data_details}. We assess sensitivity to (i) dataset scale and how it affects local anomaly contrast, (ii) imbalance between test and reference sample sizes (i.e., $p=n_\mathcal{Y}/(n_\mathcal{X}+n_\mathcal{Y})\neq 0.5$), and (iii) anomaly morphology and multiplicity. We also report the impact of varying the maximum neighbourhood rank $K_M$ and repeat each configuration across multiple random seeds; figures report means with $3\sigma$ intervals.  For this sensitivity study we consider a 10-dimensional Gaussian background with mixed-shape oversensitivity anomalies as detailed below. We defer the assessment of false-positive control under null comparisons to \ref{sec:SI1g_manifold}.

Together, the following tests display how \EagleEye is stable under substantial variations in dataset scale, sample-size imbalance, and anomaly morphology: performance degrades primarily when local anomaly contrast is diluted, and remains stable when global background structure changes away from the anomalies.  

\SIsubsection{Dataset construction}
\label{sec:toriodal_dataset}
For this section we construct new synthetic data. Reference and test samples are generated in $\mathbb{R}^{10}$ with an isotropic Gaussian background. The test sample is perturbed by injected anomalies of two morphologies at characteristic scale $\sigma_a=0.3$:
\begin{itemize}
\item \textbf{Torus anomaly.} For each injected point, the first three coordinates are drawn uniformly from a torus with major radius $R_a=\sigma_a$ and minor radius $r_a=R_a/6$, while the remaining seven coordinates are sampled from $\mathcal{N}(0,\sigma_a)$.
\item \textbf{Gaussian anomaly.} All ten coordinates are sampled from $\mathcal{N}(0,\sigma_a)$.
\end{itemize}
Unless stated otherwise, we use $K_M=500$, $q=10^{-2}$, and calibrate the null threshold using $p_{\mathrm{ext}}=10^{-5}$. For each configuration we report the cardinalities of the stagewise outputs (flagged, IDE-pruned, and rep\^{e}chage sets). When injected points are available by construction, we additionally report recovered signal counts (as in Table~\ref{tab:S1}). In all cases, results are averaged over multiple random seeds and displayed with $3\sigma$ uncertainty bands.
\begin{figure}[h!]
    \centering
    \includegraphics[width=\textwidth]{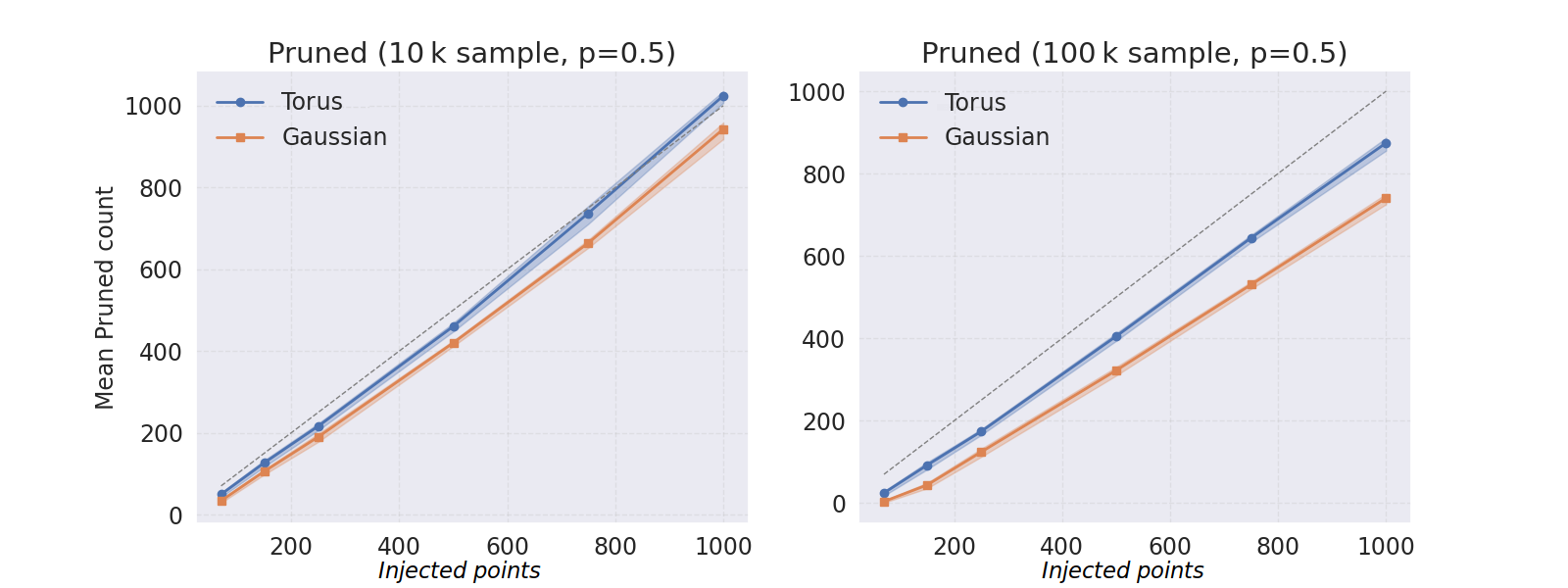} 
    \caption{\textbf{Variation in the Cardinality of the Pruned Sets as a Function of the Number of Injected Points in the Test Set}  
The plot shows the mean and the 3-sigma confidence intervals for the cardinality of the pruned sets (vertical axis) as a function of the number of injected points (horizontal axis) in the test set $\test$. Here, the maximum neighborhood rank is set to $K_M = 500$ and the default $p_\text{ext} = 10^{-5}$ is used. \textbf{Panel A:} Corresponds to datasets of 10,000 test and reference points, whereas \textbf{Panel B:} Shows datasets of 100,000 points, where the additional 90,000 points are drawn from the same underlying Gaussian distribution.}

    \label{fig:Supp_incr_card1}
\end{figure}

\SIsubsection{Increased cardinality with altered local density (dilution)}
We first increase the background sample size by an order of magnitude while holding fixed the number and location of injected anomalies. This dilutes the anomaly fraction in the test set and reduces the local enrichment of test points in neighbourhoods surrounding the anomaly. As shown in Fig.~\ref{fig:Supp_incr_card1}, this dilution decreases the size of the IDE-pruned set, reflecting reduced detection power when the local anomaly contrast is weakened. For fixed $K_M$, a point must exhibit a sufficiently large excess over the baseline neighbour proportion to be flagged (see \ref{sec:SI0b}).
\begin{figure}[h!]
    \centering
    \includegraphics[width=\textwidth]{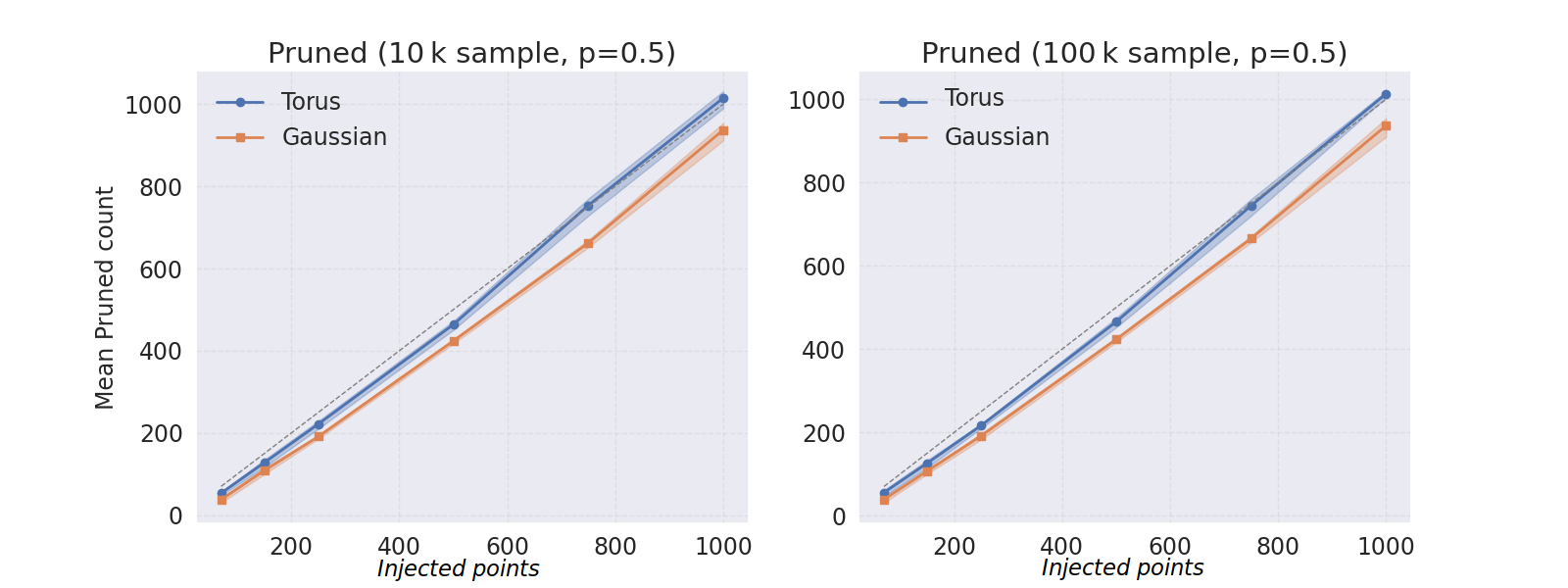} 
\caption{\textbf{Variation in the Size (Cardinality) of the Pruned Sets as a Function of the Number of Injected Points in the Test Set.}  
 The plot shows the mean and the 3-sigma confidence intervals for the cardinality of the pruned sets (vertical axis) as a function of the number of injected points (horizontal axis) in the test set $\test$. With $K_M = 500$ and the default $p_\text{ext}$, \textbf{Panel A:} Illustrates the scenario for 10,000-point datasets, while \textbf{Panel B:} Corresponds to 100,000-point datasets. In this case, the additional 90,000 points are now drawn from \textit{six} Gaussian modes arranged around the primary mode, forming a petal-like structure. }

    \label{fig:Supp_incr_card2}
\end{figure}

\SIsubsection{Increased cardinality with unaltered local density}
In a second scenario, we increase total cardinality by augmenting the background with additional points sampled from a separate Gaussian mixture arranged away from the anomaly regions. This changes overall dataset size and global background morphology while preserving the local background density in the vicinity of the anomalies. In this regime, detection performance remains stable, as shown in Fig.~\ref{fig:Supp_incr_card2}, consistent with the local nature of the underlying test statistic.

\begin{figure}[h!]
    \centering
    \includegraphics[width=\textwidth]{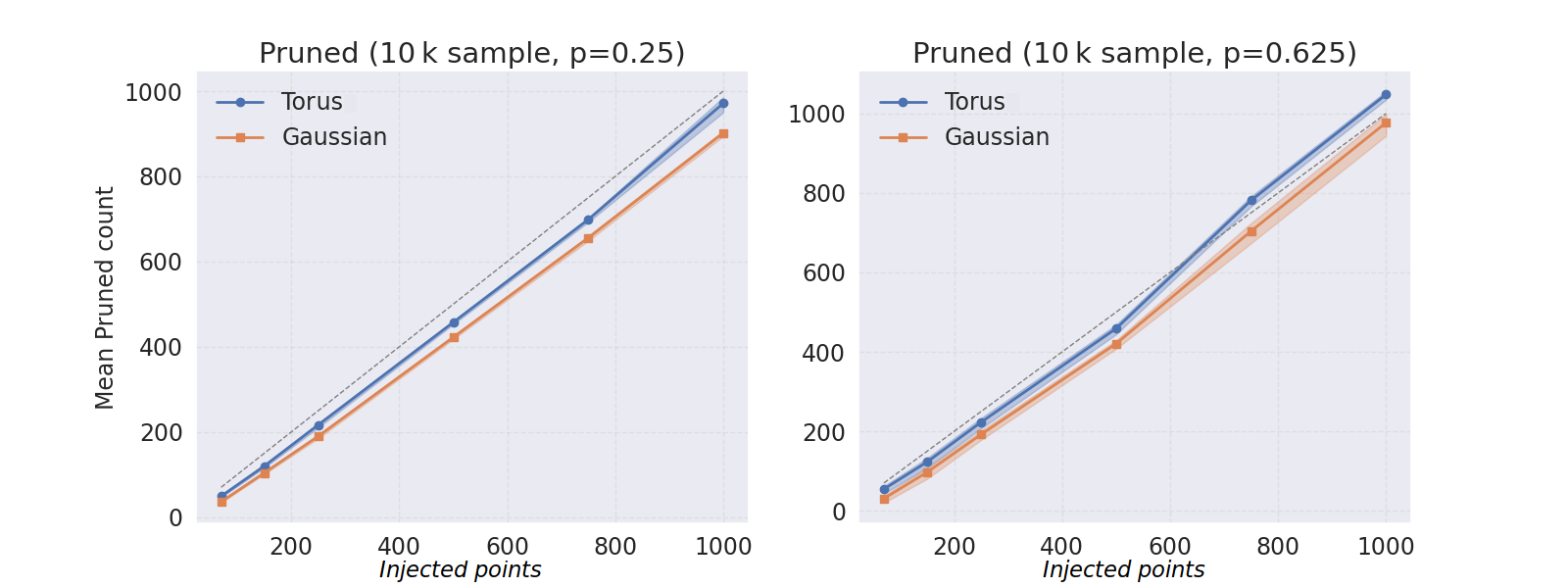} 
    \caption{\textbf{Variation in the Cardinality of Pruned Sets as a Function of the Number of Injected Points in the Test Set.}  
The plot shows the mean and the 3-sigma confidence intervals for the cardinality of the pruned sets (vertical axis) as a function of the number of injected points (horizontal axis) in the test set $\test$.In both panels, the maximum neighborhood rank is set to $K_M = 500$ and the default external probability threshold $p_\text{ext} = 10^{-5}$ is used.
 \textbf{Panel A:} represents a scenario where the test dataset consists of 10,000 points and the reference set has 30,000 points.
\textbf{Panel B:} shows the same test dataset, but in this case, the reference dataset contains only 6,000 points. }
    \label{fig:Supp_imbalance}
\end{figure}

\SIsubsection{Dataset imbalance ($p\neq 0.5$)}
Next, we assess stability when the test and reference samples have different cardinalities, so that the baseline neighbour proportion is $p=n_\mathcal{Y}/(n_\mathcal{X}+n_\mathcal{Y})\neq 0.5$. We repeat the dilution experiment of Fig.~\ref{fig:Supp_incr_card1} with a fixed test sample size and adjust the reference size to obtain $p=0.25$ and $p=0.625$. As shown in Fig.~\ref{fig:Supp_imbalance}, detection remains stable across these imbalances: \EagleEye\ continues to identify the anomalous regions and produces comparable stagewise outputs after calibration with the appropriate $p$.

\SIsection{Additional Benchmarking: Performance in high dimensions, false positive control, hyperparameter sensitivity and compute. }
\label{sec:SI1g_manifold}
\begin{figure*}[ht]
    \centering
    \includegraphics[width=0.7\linewidth]{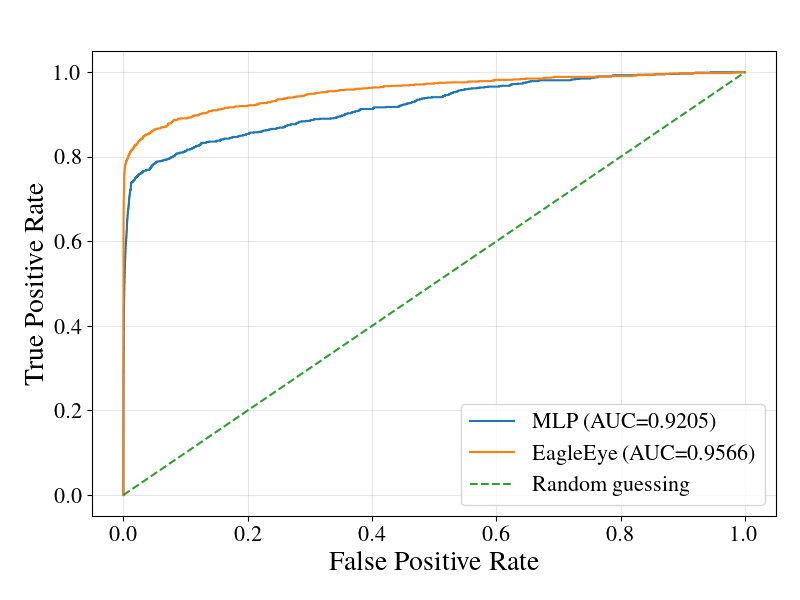}
    \caption{ Comparison for the 10D manifold in 100D of \ref{sec:SI1g_manifold}. Displayed are ROC curves for the (a) MLP anomaly score in blue and (b) \EagleEye anomaly score in orange. We emphasize that  \EagleEye then further sharpens the anomaly purity via the IDE and the Repêchage steps (see Materials and Methods). }
    \label{fig:EE_comp_MLP_100D}
\end{figure*}

In this section we provide extra benchmarking studies in order to demonstrate \EagleEye's efficacy on more demanding data sets. Specifically, we probe \EagleEye\ in a substantially more challenging regime than the toroidal morphology of \ref{sec:toriodal_dataset}. Here, the data lie on a nonlinear manifold of intrinsic dimension 10 embedded in an ambient 100-dimensional space. This construction serves two purposes: (i) to assess \textit{false-positive control} under a background-only (null) comparison, and (ii) to quantify recovery when a small fraction of injected signal is present. Unless stated otherwise, we use $q=10^{-2}$, calibrate the flagging threshold at $p_{\mathrm{ext}}=10^{-5}$ (see \ref{sec:SI1c}) , and report means and standard deviations over $N_{\mathrm{runs}}=10$ random-seed repetitions.

For each dataset, we draw latent coordinates $z=(z_0,\ldots,z_9)$ with
\[
z_j \sim \mathrm{Unif}[-10,10], \qquad j=0,\ldots,9,
\]
and embed them in $\mathbb{R}^{100}$ by padding with zeros: $(z_0,\ldots,z_9,0,\ldots,0)$. We then apply a nonlinear mapping $f:\mathbb{R}^{10}\to\mathbb{R}^{10}$ componentwise:
\[
\begin{aligned}
f_{0} &= \tfrac{2}{\pi}\arctan(z_{0}+z_{1}), &
f_{1} &= \tanh(0.5\,z_{2}), \\
f_{2} &= \tfrac{2}{1+e^{-(z_{3}+z_{4})}}-1, &
f_{3} &= \tfrac{z_{5}}{1+|z_{5}|}, \\
f_{4} &= \exp\!\left[-(z_{6}-z_{7})^{2}/2\right], &
f_{5} &= \cos(W z + 2\pi\,\xi), \\
f_{6} &= \tfrac{z_{8}z_{9}}{1+|z_{8}z_{9}|}, &
f_{7} &= \tfrac{\|(z_{0},z_{1})\|}{1+\|(z_{0},z_{1})\|}, \\
f_{8} &= \tfrac{\log(1+|z_{2}|)}{1+\log(1+|z_{2}|)}, &
f_{9} &= \mathrm{sign}(z_{3})\,\tfrac{\sqrt{|z_{3}|}}{1+\sqrt{|z_{3}|}}.
\end{aligned}
\]
Here $W\in\mathbb{R}^{10\times10}$ is a random weight matrix, and $\xi\sim\mathrm{Unif}[0,1]$ is drawn i.i.d.\ per point. Finally, we place the manifold in a generic orientation by applying a random $100\times 100$ orthogonal matrix $R$:
\[
x \;=\; R\,(f(z),0,\ldots,0)\in\mathbb{R}^{100}.
\]
For each run, $W$ and $R$ are drawn once and then \emph{shared} between the reference and test samples, so that $\mathcal{X}$ and $\mathcal{Y}$ differ only through sampling variability (null) or signal injection (alternative). The random seed controls all sources of randomness. Signal points are generated by drawing latent coordinates from a correlated Gaussian,
\[
z \sim \mathcal{N}(0,\Sigma),
\]
with $\Sigma_{ii}=1$, $\Sigma_{01}=\Sigma_{10}=0.8$, $\Sigma_{23}=\Sigma_{32}=0.9$, $\Sigma_{57}=\Sigma_{75}=-0.7$, and all other covariances equal to zero, followed by the same nonlinear map $f$ and the same rotation $R$ used for the background. Unless stated otherwise, we inject a fraction $1\%$ of such signal points into the test sample $\mathcal{Y}$, while the reference sample $\mathcal{X}$ contains background only.

\SIsubsection{Null consistency (false-positive control).}
We compare two independent background-only datasets ($\mathcal{X}$ and $\mathcal{Y}$ constructed as above with no injected signal), each of size $N=100{,}000$, using $K_M=500$ and $q=10^{-2}$. Under the null, the expected number of flagged points is $Np_{\mathrm{ext}}\approx 1$ for $p_{\mathrm{ext}}=10^{-5}$. Across $N_{\mathrm{runs}}=10$ seed repetitions, the observed maximum number of flagged points per dataset is $\le 1$, with mean $0.4$ and standard deviation $0.49$ (consistent across the flagged, IDE-pruned, and rep\^{e}chage stages). This demonstrates specificity in a high-dimensional setting where the data occupy a nonlinear manifold.

\SIsubsection{Recovery with injected anomalies}
We repeat the same setup but inject $1\%$ signal into the test sample. Over $N_{\mathrm{runs}}=10$ repetitions, \EagleEye\ reliably detects the presence of anomalies (mean flagged $626.9\pm 21.3$; mean IDE-pruned $541.1\pm 22.8$; mean rep\^{e}chage $617.3\pm 21.3$). Relative to the $1{,}000$ injected signal points, the rep\^{e}chage stage recovers on average more than $60\%$ of injected signal points. As expected for a density-based method when background and signal supports overlap, recovery is not perfect: a subset of injected points are locally indistinguishable from background and cannot be separated without incurring excessive false positives.

\SIsubsection{Effect of increasing $\mathbf{K_M}$}
We also repeat the injected-signal experiment with $K_M=1000$ (all other settings unchanged). Detection remains stable (mean flagged $648.4\pm 22.7$; mean IDE-pruned $553.4\pm 23.1$; mean rep\^{e}chage $641.2\pm 23.3$).

\SIsubsection{Computation times}
Wall-clock runtimes are reported for our current implementation on the following system: 32\,GB RAM; Intel\textsuperscript{\textregistered} Xeon\textsuperscript{\textregistered} W-2245 @ 3.90\,GHz (16 cores); NVIDIA Quadro RTX 4000. For $N=100{,}000$ points per dataset in 100D, the average end-to-end runtime is $77.84$\,s with $K_M=500$, of which the initial nearest-neighbour search accounts for $91.3\%$ of total time. Increasing to $K_M=1000$ raises the runtime to $129.44$\,s, with the nearest-neighbour step remaining the dominant cost. Runtime scales with dataset size in the expected manner for neighbour-search--dominated workloads (e.g., $215.15$\,s for $N=200{,}000$ and $27.04$\,s for $N=50{,}000$ at $K_M=500$), while memory usage remained stable in these runs.

\SIsection{Comparisons with commonly used two-sample classifiers} 
\label{sec:two_sample_compt_intro}

Two-sample anomaly detection in high-dimensional scientific datasets is commonly
approached using classifier-based likelihood-ratio estimators or transform-based
tests \cite{cranmer2015approximating,sugiyama2012density,durkan2020contrastive,thomas2022likelihood}.
However, such methods face intrinsic difficulties when anomalies are very
localised, namely they occupy a small regions of feature space. In these regimes,
classifier gradients vanish, wavelet coefficients disperse anomalous support across
many scales, and statistical calibration typically requires extensive resampling. By contrast, \EagleEye provides a \emph{pointwise}, \emph{calibrated} hypothesis test based on Bernoulli and binomial statistics, yielding a closed-form null distribution
and an analytic significance threshold without classifier training, density modelling,
or resampling (see main text). To illustrate these distinctions, we compare \EagleEye with two representative
approaches: (i) a multilayer perceptron (MLP) classifier and (ii) a wavelet-based
two-sample test. Each comparison uses a synthetic dataset designed to expose known
failure modes of the corresponding method. 
Code reproducing these comparisons is
available at \url{https://github.com/sspring137/EagleEye/tree/main/examples/2sample_comparison}.

\SIsubsection{MLP classifier vs.\EagleEye on the 100D example of Sec. \ref{sec:SI1g_manifold}}

Optimising a binary cross-entropy to train a MLP to distinguish a reference
(background) sample from a test (data) sample yields, in the large-sample/large-capacity
limit, the Bayes-optimal classifier
\[
f^\star(x)=\Pr(Y=1\mid x)
=\frac{\pi_1\,p_{\mathrm{data}}(x)}{\pi_0\,p_{\mathrm{bg}}(x)+\pi_1\,p_{\mathrm{data}}(x)}\,,
\]
where $\pi_0,\pi_1$ are the class priors \cite{cranmer2015approximating}.
With equal priors, the classifier odds
\[
\widehat r(x)\;\equiv\;\frac{f^\star(x)}{1-f^\star(x)}
\]
recover the density ratio between the two samples,
\[
\widehat r(x)=\frac{p_{\mathrm{data}}(x)}{p_{\mathrm{bg}}(x)}\,.
\]
We  use $\widehat r(x)$ as the MLP-derived pointwise anomaly score for reasons that follow: Assume the test distribution is a mixture
\[
p_{\mathrm{data}}(x)=(1-\varepsilon)\,p_{\mathrm{bg}}(x)+\varepsilon\,p_{\mathrm{sig}}(x).
\]
Then
\[
\frac{p_{\mathrm{data}}(x)}{p_{\mathrm{bg}}(x)}
=1-\varepsilon+\varepsilon\,\frac{p_{\mathrm{sig}}(x)}{p_{\mathrm{bg}}(x)},
\]
so the two-sample density ratio is a strictly monotonic function of the
signal--background likelihood ratio wherever $p_{\mathrm{bg}}$ has support.
Equivalently, a classifier trained to separate $\test$ from $\refe$ is (up to a
monotonic rescaling) also optimal for separating signal from background, as formalised
by Theorem~1 of the CWoLa framework \cite{metodiev2017classifier}.

In Fig.~\ref{fig:EE_comp_MLP_100D} we compute the anomaly score $\hat{r}$ for the 10D manifold defined at the beginning of this section and compare its classification performance to that of \EagleEye. The resulting curves show that \EagleEye outperforms the MLP already at the first stage of the pipeline.

\SIsubsection{MLP classifier vs.\ \EagleEye on a 20D ``needle-in-a-haystack'' anomaly}
\label{sec:mlp_comp_studies}
    \begin{figure*}[t]
    \centering
    \includegraphics[width=0.7\linewidth]{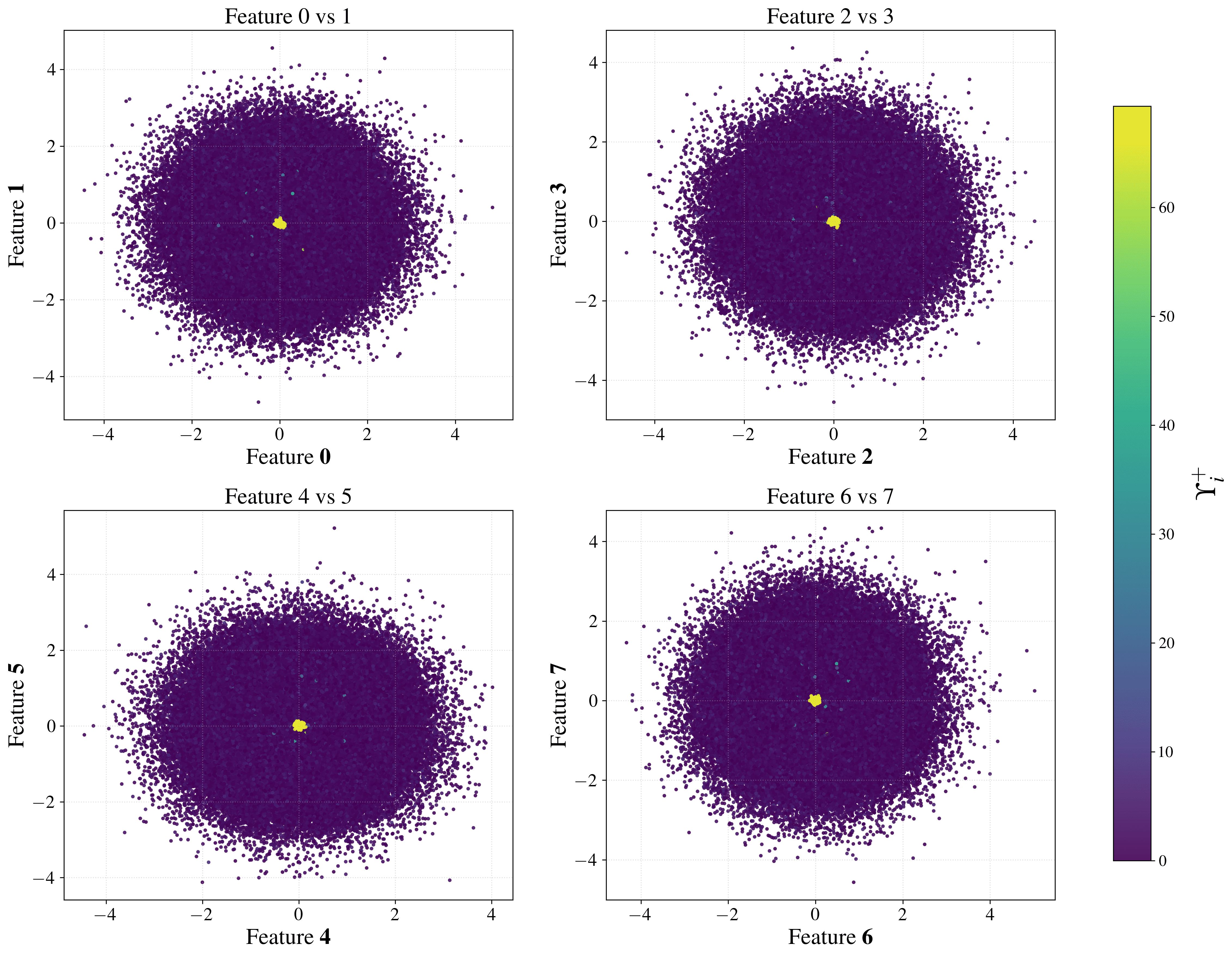}
    \includegraphics[width=0.7\linewidth]{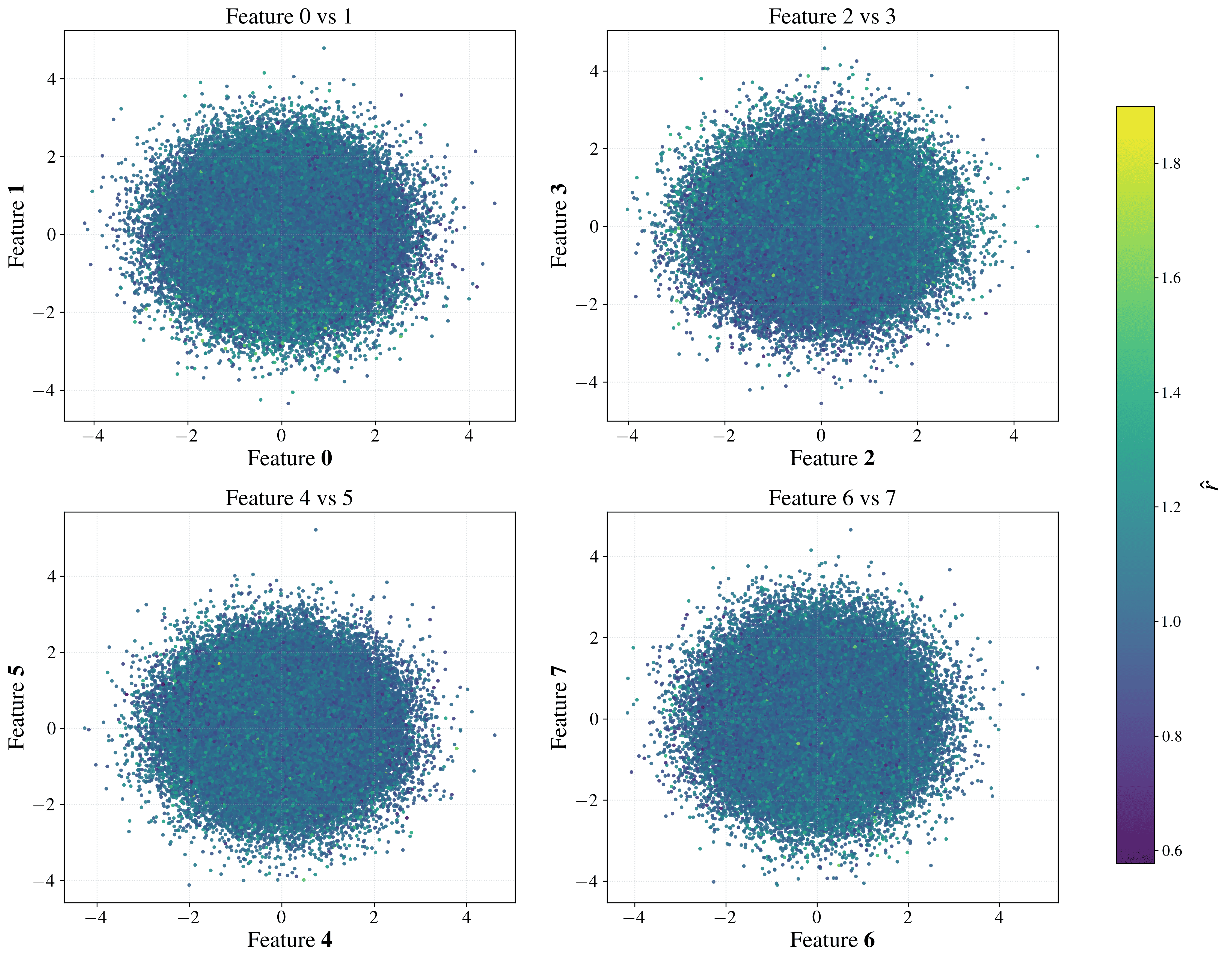}
    \caption{\textbf{Looking for a needle in a haystack. Top:} Marginal scatter plots in some projections of the 20 dimensions used to demonstrate \EagleEye's performance in identifying a tightly concentrated anomaly consisting of 100 points in a background of $2\times10^5$ points sampled from a uniform Gaussian. The colour bar displays the anomaly score $\Upsilon^+_i$ for each point. \textbf{Bottom:} The same as the top figure but displaying the likelihood ratio estimate $\hat{r}$ obtained with the MLP classifier. }
    \label{fig:MLP_comp_MLP}
\end{figure*}
Here we consider an extreme anomaly-detection benchmark in which the reference sample is drawn from
$p_{\mathrm{bkg}}(x)=\mathcal{N}(0,I_{20})$ with $n_{\mathrm{bkg}}=2\times10^5$,
and the test set contains the same background plus $n_{\mathrm{sig}}=100$ anomalous points drawn from a sharply concentrated Gaussian, $p_{\mathrm{sig}}(x)=\mathcal{N}(0,\sigma^2 I_{20})$ with $\sigma=0.05$.
Although the fraction of signal points in the dataset is small but non-negligible,
$\varepsilon = n_{\mathrm{sig}}/n_{\mathrm{tot}}\sim5\times10^{-4}$, the signal occupies an exponentially small region of the 20-dimensional feature space. Specifically, the probability mass of a signal-scale neighbourhood under the background distribution scales as $\mathbb{P}_{\mathrm{bkg}}(\|x\|\lesssim\sigma)\sim\sigma^{20}\approx10^{-26}$.

Results are shown in Fig.~\ref{fig:MLP_comp_MLP}.
We observe stochastic-gradient training of an MLP failing catastrophically in this regime, with none of the injected signal points assigned large likelihood-ratio scores $\hat r$. For a minibatch of size $B$, the probability that at least one signal point is sampled is $P\simeq\varepsilon B$, so signal examples are occasionally seen during training.
However, the expected gradient contribution associated with parameters sensitive only to the localised signal region is suppressed by $\mathcal{O}(\epsilon \sigma^d )$. Consequently, even when signal points are sampled, the corresponding gradient updates are negligible compared to those driven by the background. 


We note that the  "gradient dilution'' detailed above is specific to regimes in which the anomaly fraction is extremely small. If the background sample size is reduced while keeping the number of signal points fixed, the probability volume of the signal region under the background distribution remains unchanged, but the increased contamination fraction ensures that signal points are sampled more frequently during training, allowing classifier-based methods to potentially recover the anomaly. This is further exacerbated in lower dimensions where the point becomes moot as the anomaly detection task becomes trivial. An extensive comparison of the performance of an MLP vs \EagleEye, and indeed nearest neighbour methods in general, in varying anomaly purity contexts remains interesting but was deemed beyond the scope of the current work. 

The reported results are obtained using an MLP with three hidden layers of width 64, trained for up to 100 epochs with early stopping, a learning rate of $10^{-3}$, and the Adam optimizer with minibatch size $B=128$.
We observe qualitatively similar behaviour under variations of these hyperparameters.
This example highlights that, in anomaly-detection settings, care must be taken when relying on classifier-based methods, as the onset of gradient suppression due to extreme localisation is not always apparent from dataset-level class imbalance alone.

\EagleEye replaces gradient-based learning with an analytic, pointwise hypothesis test based on local event counts. With $p_{\mathrm{ext}}=10^{-5}$, \EagleEye obtains a threshold $\Upsilon_+^\star\approx13.6$ and flags all 100 anomalous points. After iterative density equalisation, a compact cluster remains, and the method achieves AUC $=1$ (varying $\Upsilon_+^\star=13.6$) . This example highlights \EagleEye’s robustness in regimes where classifier-based two-sample tests fail due to vanishing gradients, illustrating the advantage of its analytic Bernoulli/binomial hypothesis-testing framework. We display the results in the same four demonstrative figures in Fig~\ref{fig:MLP_comp_MLP}. We also emphasize that this result is obtained after a single call to \EagleEye with conservative hyperparameter choices $K_M=200$ and $p_\text{ext} = 10^{-5}$.

\SIsubsection{Kernel-based classifier vs.\ \EagleEye on a 3D spiral anomaly}
\label{sec:kernal_based}
We next consider a qualitatively different anomaly: a high-density but highly structured
signal with nontrivial morphology. The background consists of $n_{\mathrm{bg}}=10^4$ points uniformly distributed
in the cube $[-1,1]^3$, while the signal comprises $n_{\mathrm{sig}}=5\times10^3$ points
arranged along a thin three-dimensional spiral with small Gaussian width.
Unlike the ``needle-in-a-haystack'' example of Sec.~\ref{sec:mlp_comp_studies}, the signal here is not rare
in abundance but occupies a geometrically complex, curved manifold embedded in the space.

\begin{figure*}[t]
    \centering
        \includegraphics[width=0.5\linewidth]{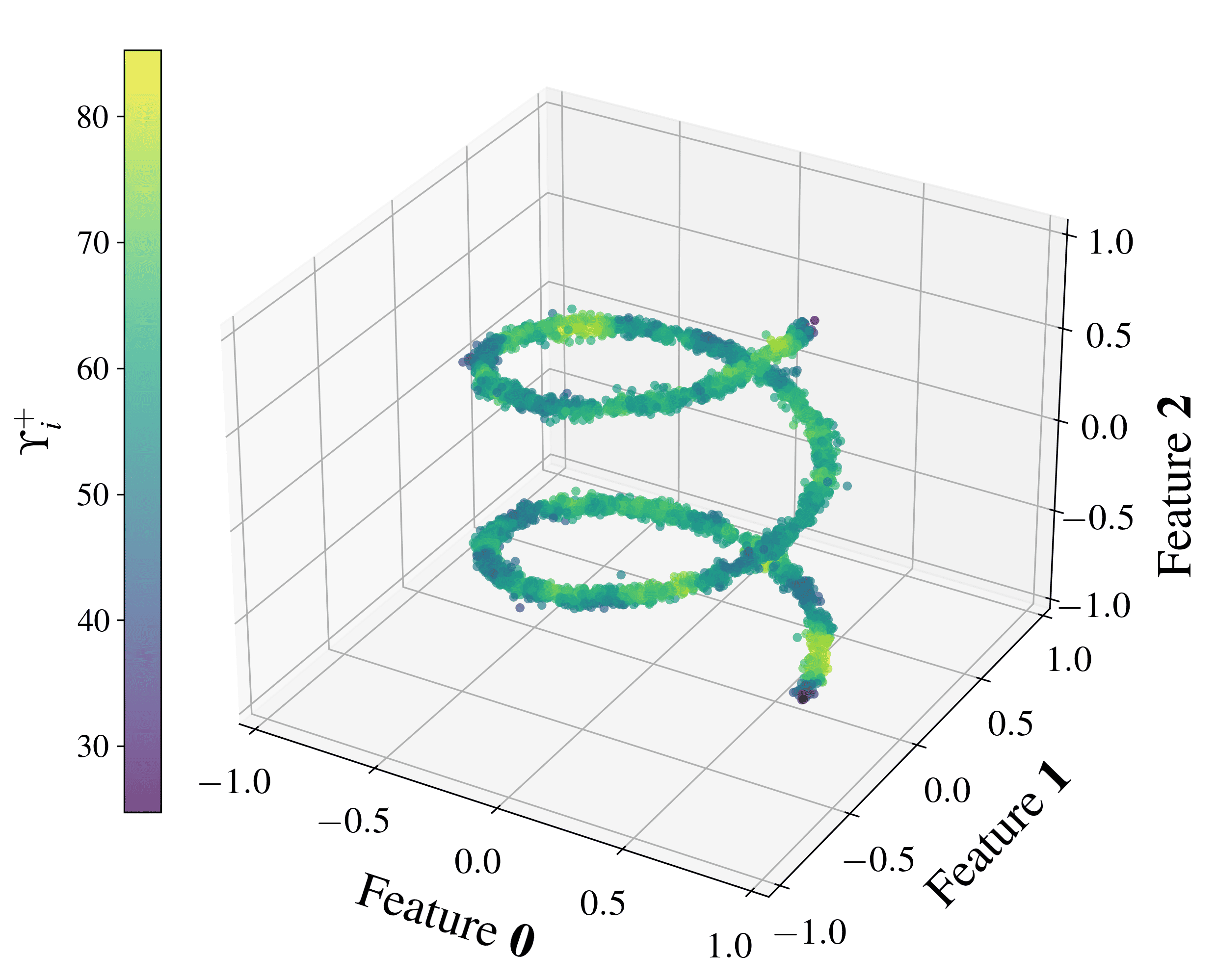}~\includegraphics[width=0.5\linewidth]{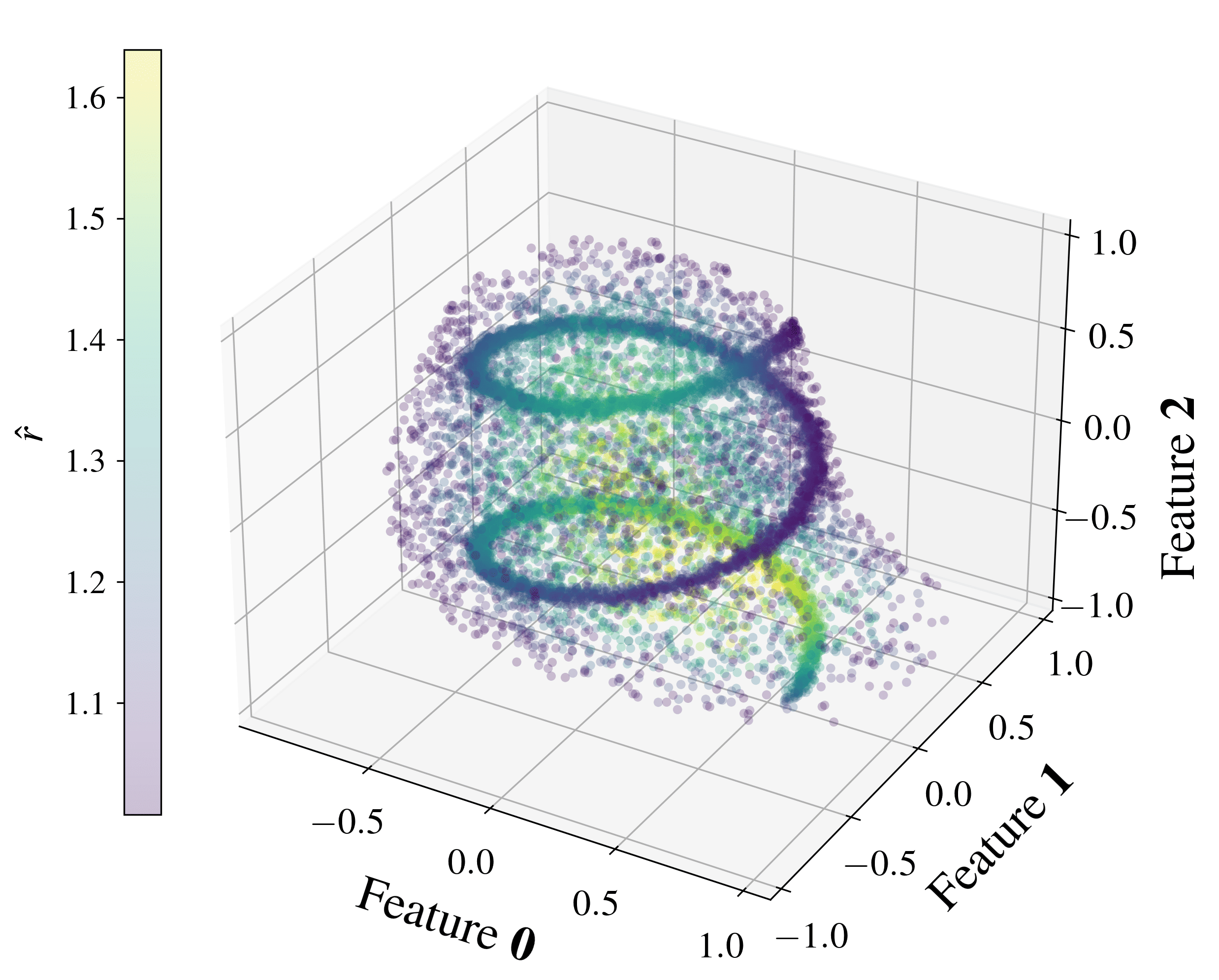}
\caption{
\textbf{Pointwise anomaly scores for \EagleEye and an RBF-kernel classifier on the 3D spiral dataset.}
\textbf{Left:} \EagleEye scores $\Upsilon_i^{+}$ shown for the IDE-pruned set only. All displayed points satisfy the statistically defined threshold $\Upsilon_i^{+}\geq \Upsilon_{+}^{\ast} \simeq 24.8$.
\textbf{Right:} RBF-kernel classifier likelihood-ratio estimate $\hat r(x)$ for the same data after transforming $\phi(x)\rightarrow x$.
Since $\hat r$ has no associated null distribution or calibrated threshold, we adopt the  ad hoc visual cut $\hat r>1$.
}
    \label{fig:kernal_comp}
\end{figure*}

As a representative kernel-based two-sample classifier, we use an explicit
random Fourier feature approximation to the RBF kernel, followed by a linear logistic classifier as detailed in \cite{rahimi2008weighted} and as implemeted in the \texttt{Scikit Learn} library \cite{scikit-learn}.
Each point $x$ is mapped to a high-dimensional feature vector $\phi(x)$ such that
$\langle \phi(x),\phi(x')\rangle \approx \exp(-\gamma\|x-x'\|^2)$.
The kernel bandwidth $\gamma$ is set heuristically to the inverse squared median
pairwise distance estimated on a $10^4$-point subsample, and the feature map uses
$4000$ random components. A linear logistic classifier is then trained in this feature space to discriminate the sets $\refe$ vs. $\test$ (where $\test$ contains the spiral signal). The output of the classifier is used to obtain a $\hat{r}(\phi(x))$ and again serves as the anomaly score.  We display the results in the right plot of Fig.~\ref{fig:kernal_comp}. The kernel classifier assigns elevated scores to broad regions surrounding the spiral but fails to localise the anomaly sharply.
Because the RBF kernel responds isotropically in Euclidean distance, the spiral
structure is smeared across many kernel components, diluting contrast between signal and background. Moreover, the resulting anomaly score $\hat{r}(\phi(x))$, as was the case for the MLP in Sec. SI3a,  lacks a closed-form null distribution, so thresholds must be chosen heuristically or via resampling. Regardless, we observe a ROC AUC $\simeq 0.75$ obtained after varying the threshold. We observe similar behaviour after varying the feature map dimensionality within the range $[3, 4\times10^{4}]$ as well as the subsampling partition size. We note the extreme memory usage of the wavelet classifier which exceeded $\simeq 50$ Gb of RAM even though the total data size was $\mathcal{O}(10^4)$. 

By contrast, \EagleEye detects the spiral as a single, well-defined overdensity
without any kernel tuning or training as is shown in the left plot of Fig.~\ref{fig:kernal_comp}, where we display only the pruned set $\hat{\test}$ obtained after IDE with $p_{\mathrm{ext}}=10^{-5}$ and $K_M=200$, and therefore represents the excess 'mass' contained within the spiral. \EagleEye assigns  calibrated pointwise anomaly scores, recovers the full spiral morphology, and cleanly separates signal from background with a ROC AUC = $0.99$. This example highlights that kernel-based classifiers, while flexible, can struggle with geometrically complex anomalies and lack statistically grounded thresholds, whereas \EagleEye remains both sensitive and interpretable.

This manifold-based stress test shows that \EagleEye\ (i) maintains false-positive control under null comparisons at the nominal $p_{\mathrm{ext}}$ level, and (ii) retains substantial anomaly-recovery power when signal is injected, even when the data lie on a nonlinear low-dimensional manifold embedded in a high-dimensional ambient space. Computationally, the dominant cost arises from nearest-neighbour search.

\SIsection{Supporting information for the collider physics application}
\label{sec:SI2_LHC}
In this section we elaborate on the study shown in Results Sec.~\ref{sec:LHC_results} of the main text, specifically pertaining to feature selection/engineering and details regarding the background estimates $\refe_1$ and $\refe_2$. We additionally provide complimentary results to those presented in Fig.~\ref{fig:LHC_locations}. 
\SIsubsection{Dataset details and feature engineering}
\label{sec:lhc_features}
The Large Hadron Collider (LHC) \cite{Lyndon_Evans_2008} at CERN in Geneva, Switzerland, is one of the most ambitious scientific instruments ever built, responsible for the discovery of the Higgs boson in 2012 by the ATLAS and CMS collaborations \cite{Higgs1964,ATLAS2012,CMS2012}. Searches for physics beyond the Standard Model (BSM) increasingly rely on anomaly–detection strategies capable of identifying small deviations from an expected background distribution. A central challenge is that the background is never uniquely known: different Monte Carlo generators, detector simulations, or nuisance-parameter variations give rise to distinct background models whose discrepancies constitute a dominant source of systematic uncertainty. In current LHC practice, such systematics are typically handled in three main ways:(1) introducing a nuisance parameter controlling shape and/or normalisation variations in one or two key observables ~\cite{CRANMER_2006};  
(2) quantifying “MC closure” residuals between nominal and alternative histograms in a small number of features \cite{CMScollaboration_2011,Fleck_2013};  
(3) performing model informed or weakly-supervised sideband-based interpolation procedures \cite{Cowan:2010js,cathode2022,CWoLa}. While powerful, these techniques are intrinsically low-dimensional and rarely provide a fully multivariate map of background disagreement. \EagleEye offers an alternative: a statistically principled, fully high-dimensional comparison of background models. First,  we construct a map of \emph{systematics regions} in feature space by contrasting two background models. This identifies regions where systematic mismodelling dominates and regions where both backgrounds agree and can therefore be trusted. Second, we identify \emph{persistent} density anomalies by requiring that they survive independently in both comparisons, $\refe_1$ vs.\ $\test$ and $\refe_2$ vs.\ $\test$, ensuring that they cannot be attributed to generator-specific mismodelling.

The task is resonant anomaly detection~\cite{dijet_intro_atlas}, targeting short-lived particles whose hadronic decays produce two jets. As a representative testbed, we use the LHC Olympics R\&D dataset~\cite{zenodo_rnd}, which provides $10^6$ high-fidelity QCD dijet events produced with \textsc{Pythia} 8 \cite{pythia} and propagated through a detector volume using \textsc{Delphes} 3.4.1 \cite{delphes}, as well as $10^5$ signal events from a heavy resonance whose decay products hadronize into jets. We also take an additional $\sim6\times10^5$ QCD background events from \cite{zenodo_additionalQCD}, generated under the same simulation specifications. The jets are then clustered using \textsc{FastJet} \cite{cacciari2006fastjetcodefastkt} using specifications as detailed in \cite{zenodo_rnd}.  Our features are based on observables constructed by the two highest- $p_T$ jets. The two jets are sorted by their invariant mass, such that $m_{J_1}<m_{J_2}$. The total feature vector is therefore: 
\begin{equation}
    d = (m_{j1},\ \Delta m_J = m_{J2} - m_{j1},\ \tau_{21}^{J_1},\ \tau_{21}^{J_2}),
    \label{eqn:LHC_features}
\end{equation}
where the $n$ subjettiness ratios are defined as $\tau_{i j} \equiv \tau_i / \tau_j$~\cite{Thaler2011}. These features were seen to capture the dominant discriminating information while avoiding redundant kinematic degrees of freedom \cite{cathode2022,Kasieczka_2021}.

\begin{figure}[t]
  \centering
  \includegraphics[width=1\textwidth]{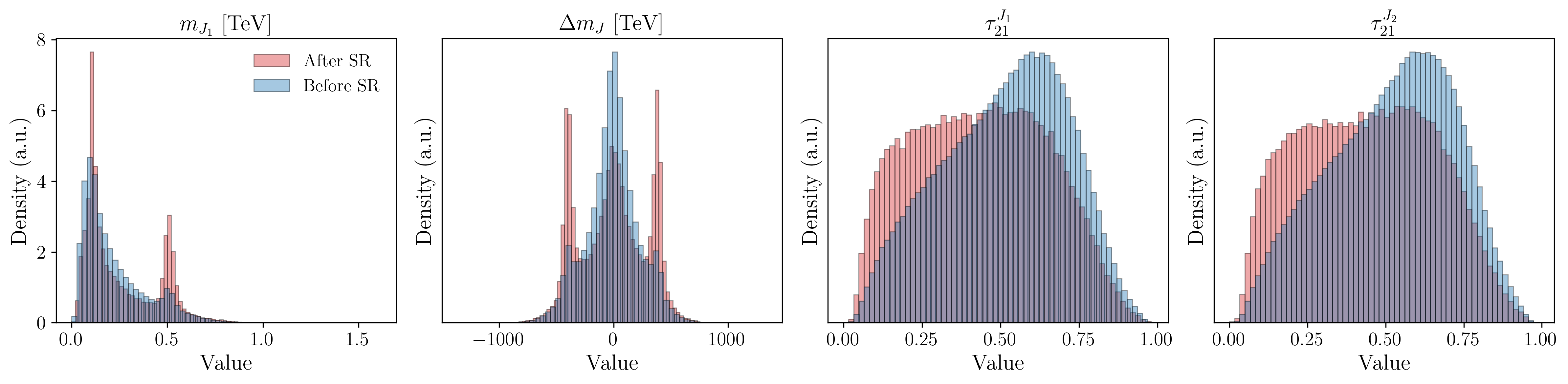}
    \includegraphics[width=1\textwidth]{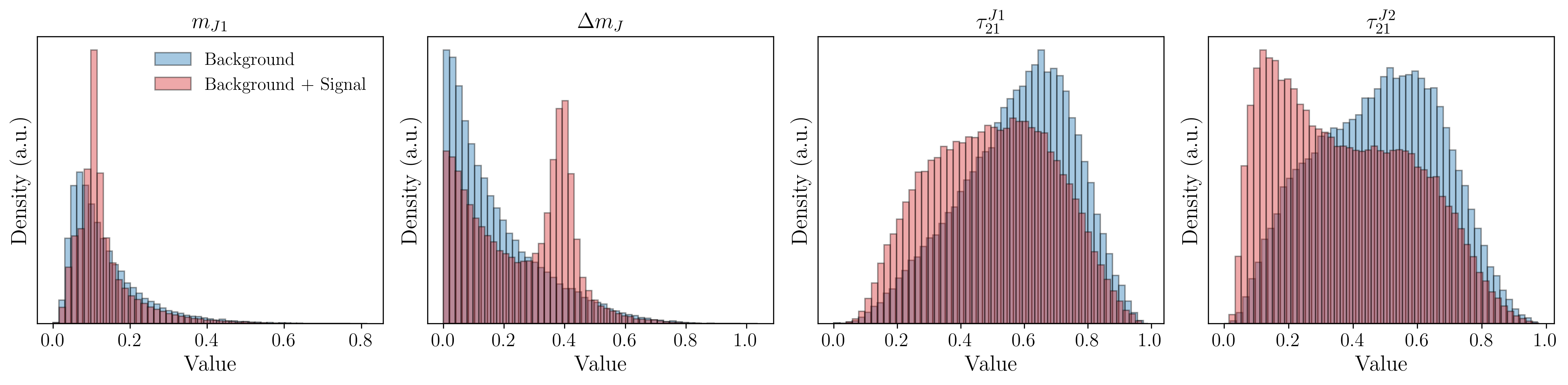}
  \caption{One dimensional marginal distributions of the features used to construct the SR and enhance the resonant density anomaly in the LHCO R\&D dataset. \textbf{Top:} Before and after the application of the SR window $m_{JJ}\in[3.3,3.7]\;\mathrm{TeV}$ in a test set comprising both signal and background (shown with a large injected signal for visibility). This single cut immediately concentrates the anomalous events, boosting their local density, manifesting as peaks in the mass distributions, and extended distributional shifts in the subjettiness parameters.
\textbf{Bottom:} Implementation of the jet-mass ordering ($m_{j2}>m_{j1}$) and rescaling both mass coordinates to the interval (0,1) TeV for two samples now consisting of pure QCD background and background + signal. These additional steps sharpen the separation between the mixed (anomolous) and pure-background samples and transforms the bimodal excess into a unimodal peak.}
  \label{fig:Scaling}
\end{figure}

Following the standard procedure for a resonance anomaly search, we conduct additional feature engineering by defining a 'signal region' (SR). A SR is a narrowly defined interval of a discriminating observable in which an excess of events from a hypothesised particle is expected to appear above the Standard-Model background.  It is chosen to maximise the signal-to-background ratio while retaining the bulk of the putative signal. The resonant feature for the LHCO is the invariant dijet mass \cite{dijet_atlas_1,dijetmass_cms}:
\begin{equation}
  m_{JJ}
  =\sqrt{\bigl(E_{J_1}+E_{J_2}\bigr)^{2}
          -\bigl(\boldsymbol{p}_{J_1}+\boldsymbol{p}_{J_2}\bigr)^{2}},
  \label{eq:mjj}
\end{equation}
with which we adopt the same signal region as Ref.~\cite{Kasieczka_2021}: $m_{JJ}\in[3.3,3.7]\;\mathrm{TeV}$. In  practice, multiple signal regions are adopted to attempt to optimise the observation of a resonance in $m_{jj}$, and therefore the analysis should be repeated in all such windows. We note that as far as \EagleEye is concerned, such practices are simply regarded as feature engineering and the methods we present here remain quite general. The top row of Fig.~\ref{fig:Scaling} demonstrates the effect of the signal region cut $m_{JJ}\in[3.3,3.7]\;\mathrm{TeV}$ on the features ${d}$. The lower row shows the effect of the mass reordering prescription, \(m_{J_2}>m_{J_1}\), which renders the anomaly nearly unimodal in this space and amplifies its contrast inside the signal window. 

For the anomaly detection analysis take the test the set $\test$ to have a signal injection percentage of  $\sim0.6\%$ (in the SR), corresponding to 776 signal and 121352 background events after SR cuts, which roughly corresponds to the signal contamination of the original LHCO data challenge.

\SIsubsection{Background models}
\label{sec:pyth_her}
We take two independent QCD background as reference sets $R=[\refe^1,\refe^2]$. As will likely be the case in a real analysis, the samples in $\refe^1$ and $\refe^2$ should not originate from the same generating process as the background samples in $\test$. $\refe^1$ is built from $10^6$ new \texttt{Pythia} events from Ref.~\cite{zenodo_official} reconstructed with different \textsc{Delphes} parameters than our test data $\test$ and reduced to our high level features in Eqn.~\eqref{eqn:LHC_features} using \texttt{FastJet}  (v3.5.1.2, different to that used for $\test$) with anti-$k_t$ clustering and a $p_T>1.2\,\mathrm{TeV}$ jet requirement. The second reference sample, $\refe_2$, is constructed following the
\textsc{Cathode} (Classifying Anomalies THrough Outer Density Estimation) method~\cite{cathode2022}. Rather than relying on a simulated Monte Carlo background,
\textsc{Cathode} estimates the background distribution in the signal region by extrapolating a learnt 
conditional density model from data in surrounding sideband regions. This yields a background estimate that is
entirely data-driven. We use \textsc{Cathode} to extract background samples within the signal region defined
above, following the training, validation, and data-splitting prescriptions
described in Ref.~\cite{cathode2022}, for which the LHCO R\&D dataset was also used. Implementation details and dataset
configurations are provided in the public CATHODE repository
\cite{cathode_github}.

After signal region cuts we observe $\refe_1$ and $\refe_2$ as having 86366 and 200000 (then subsampled to 121352) events respectively, which when analysed with $\test$, correspond to $\phat$  fractions (Eqn.~\ref{eqn:phat}) of $\phat_{\refe_1}=0.58$ and $\phat_{\refe_2}=0.50$.

We compare $\refe_1$ to $\refe_2$ using the repêchage procedure of Sec.~\ref{sec:repechage}, yielding the $\mathcal{X}_1$– and $\mathcal{X}_2$–overdensity sets whose  density fields appear as the grey (underdensity) and red (overdensity) contours in Fig.~\ref{fig:LHC_locations}. These contours map both the \emph{location} and \emph{magnitude} of generator–detector systematics.  We quantify the extent of these systematics using our  $z$-score estimates $\Lambda_{\alpha(\mathcal{X}, \rightarrow \mathcal{Y})}$ as defined in Eq.~\ref{eqn:s/b} and Sec.~\ref{sec:injection}, where the $S_{\alpha}$ in Eq.~\ref{eqn:s/b} now is interpreted as the number of points contributing over or under-densities attributed to the systematics across all repêchaged sets  $\mathcal{X}_\alpha^{j,\text {anom}}$. Regions of feature space unaffected by systematics constitute domains in which both background models are in agreement, and form the basis for identifying persistent density anomalies in Sec.~\ref{sec:LHC_results}. We emphasize that in general it is possible that signal density anomalies lie in the same domain as a systematic over/underdensity.  While typical LHC analyses aim either to avoid such systematic-dominated regions or to constrain them with dedicated control/validation regions and nuisance-parameter fits, a full development and evaluation of methods to recover tight, signal-like anomalies that coincide with these regions is beyond the scope of this paper and will be explored in future work.

\SIsubsection{ Additional results: putative anomaly identified as systematics}
\label{sec:additional_lhc_systematic}
\begin{figure}[ht]
    \centering
    \includegraphics[width=0.7\linewidth]{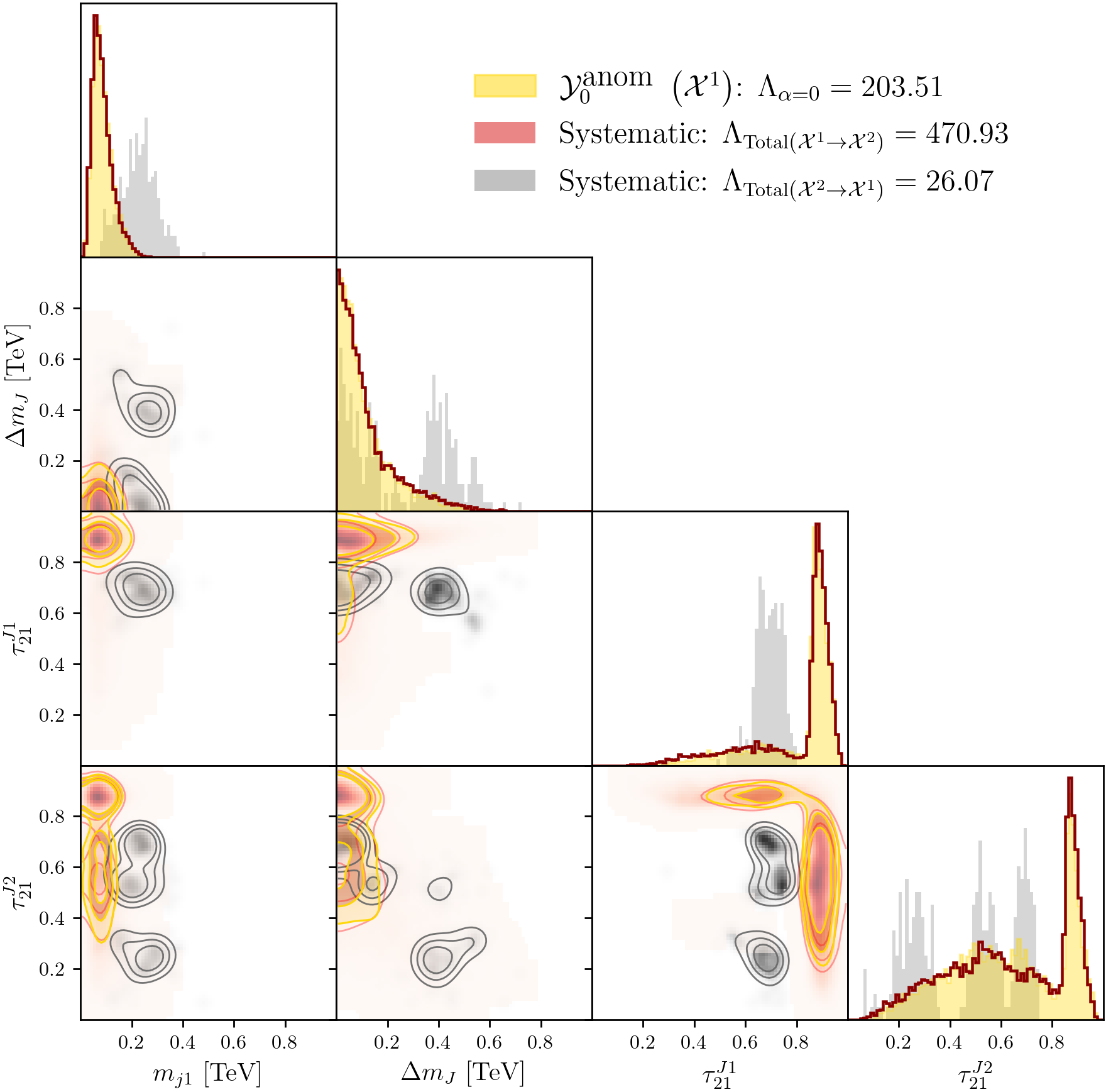}
    \caption{
\textbf{Spurious anomaly rejected by persistent anomaly search is localized within a systematics-dominated overdensity.}  
Same setup as Fig.~\ref{fig:LHC_locations}, but showing the overdensity $\mathcal{Y}^{\text{anom}}_{0}(\refe^1)$ (yellow), identified as an overdensity in $\test$ induced by the same systematic background modelling difference responsible for producing $\alpha\left(\mathcal{X}^1 \rightarrow \mathcal{X}^2\right)$ (red). While this cluster exhibits a large estimated significance (Tab.\ref{tab:LHC_cluster_summary}), it has zero overlap with clusters from the alternative background model, and it is thus discarded. 
}
    \label{fig:systematic}
\end{figure}
For completeness, Fig.~\ref{fig:systematic} displays the second anomaly cluster $\mathcal{Y}_0^{\text {anom }}\left(\refe^1\right)$, overlaid onto the same 
generator–detector systematics map shown in the main text (Fig.~\ref{fig:LHC_locations}).  As shown in Tab.~\ref{tab:LHC_cluster_summary}, this cluster contains 
$4799$ points, and displays a large local significance $\Lambda_{0\left(\mathcal{X}^1 \rightarrow \mathcal{Y}\right)}=203.51$. It is discarded by criterion (2) for a persistent anomaly as it has no overlap with the cluster $\mathcal{Y}_0^{\text {anom }}\left(\refe^2\right)$. Criterion two enforces that a persistent anomaly be present in $\test$ with respect to both reference sets $\refe^1,\refe^2$. Given that this strong over density has no counter part, it is likely induced by background mismodelling in one of the reference sets (with respect to the true background present in the test data $\test$). Indeed, inspection of the figure shows that it is located in the same region of parameter space as  the overdensity in the \textsc{Cathode} background model, $\refe_2$, when compared with the \textsc{Pythia} background, $\refe^1$ i.e $\alpha\left(\mathcal{X}^1 \rightarrow \mathcal{X}^2\right)$ (red contours).  This would imply that $\mathcal{Y}_0^{\text {anom }}\left(\refe^1\right)$ is induced by the same background miss-modelling that was responsible for the systematics $\alpha\left(\mathcal{X}^1 \rightarrow \mathcal{X}^2\right)$. We note that a quantitive assertion of this association is beyond the scope of this work, but remains interesting and of potential significance to the physics community, and is therefore left for future investigation.

\SIsubsection{Significance of persistent anomaly for the general multi-reference case}
\label{sec:multi_ref_significance} The following prescription works for any general number of observed persistent anomalies in a given test set $\test$. For simplicity, below we consider the case of only one. 

Let \(\{\refe^j\}_{j=1}^R\) be \(R\) reference sets.  Let
$\Lambda_{\alpha\left(\mathcal{X}^j \rightarrow \mathcal{Y}\right)}$ be our familiar $z$-score for cluster $\alpha$ associated with the anomaly detected against reference \(\refe^j\). Define the persistent anomaly set as the union of all repêchage clusters  that survive the persistence criterion specified in the main text Results \ref{sec:LHC_results}:
\[
\test^{\text{persist}} \;=\; \bigcup_{j=1}^R \mathcal{Y}^{\mathrm{anom}}_{\alpha(\refe^j)} .
\]
For any test point \(Y_i\in\test^{\text{persist}}\), we define the weighted  $z$-score $\bar{\Lambda}_i$, averaged over all reference sets $j$ that flagged it. 
The persistent (weighted) $z$-score of the union \(\test^{\text{persist}}\) is
then the arithmetic mean of the per-point significances,
\begin{equation}
\Lambda^{\mathrm{Persist}}
\;=\;
\frac{1}{|\test^{\text{persist}}|}\sum_{Y_i\in\test^{\text{persist}}}
\bar{\Lambda}_i.
\label{eq:lambda_persist_pointwise} \nonumber
\end{equation}
In the two-reference case \(R=2\), with anomalous sets
\(\mathcal{Y}_{1\,(\mathcal{X}^{\mathbf{1}})}^{\mathrm{anom}}\) and
\(\mathcal{Y}_{0\,(\mathcal{X}^{\mathbf{2}})}^{\mathrm{anom}}\),
and corresponding significances
\(\Lambda_{1(\refe^1\rightarrow\test)}\) and
\(\Lambda_{0(\refe^2\rightarrow\test)}\), the pointwise score
\(\Lambda(y)\) equals \(\Lambda_{1(\refe^1\rightarrow\test)}\) for
points only in \(\mathcal{Y}_{1\,(\mathcal{X}^{\mathbf{1}})}^{\mathrm{anom}}\),
\(\Lambda_{0(\refe^2\rightarrow\test)}\) for points only in
\(\mathcal{Y}_{0\,(\mathcal{X}^{\mathbf{2}})}^{\mathrm{anom}}\), and
\(\tfrac{1}{2}\bigl(\Lambda_{1(\refe^1\rightarrow\test)}+\Lambda_{0(\refe^2\rightarrow\test)}\bigr)\)
for points in the intersection. Hence the persistent significance has the closed form
\begin{equation}
\Lambda^{\mathrm{Persist}}
=
\frac{
\bigl|\mathcal{Y}_{1\,(\mathcal{X}^{\mathbf{1}})}^{\mathrm{anom}}\setminus
\mathcal{Y}_{0\,(\mathcal{X}^{\mathbf{2}})}^{\mathrm{anom}}\bigr|\,
\Lambda_{1(\refe^1\rightarrow\test)}
+
\bigl|\mathcal{Y}_{0\,(\mathcal{X}^{\mathbf{2}})}^{\mathrm{anom}}\setminus
\mathcal{Y}_{1\,(\mathcal{X}^{\mathbf{1}})}^{\mathrm{anom}}\bigr|\,
\Lambda_{0(\refe^2\rightarrow\test)}
+
\bigl|\mathcal{Y}_{1\,(\mathcal{X}^{\mathbf{1}})}^{\mathrm{anom}}\cap
\mathcal{Y}_{0\,(\mathcal{X}^{\mathbf{2}})}^{\mathrm{anom}}\bigr|\,
\frac{\Lambda_{1(\refe^1\rightarrow\test)}+\Lambda_{0(\refe^2\rightarrow\test)}}{2}
}{
\bigl|\mathcal{Y}_{1\,(\mathcal{X}^{\mathbf{1}})}^{\mathrm{anom}}
\cup
\mathcal{Y}_{0\,(\mathcal{X}^{\mathbf{2}})}^{\mathrm{anom}}\bigr|
}\!.
\label{eqn:lambda_peersist}
\end{equation}


\end{document}